\documentclass{article}

\usepackage{microtype}
\usepackage{graphicx}
\usepackage{subfigure}
\usepackage{booktabs} 

\usepackage{hyperref}

\usepackage[accepted]{icml2023}

\usepackage{amsmath}
\usepackage{amssymb}
\usepackage{mathtools}
\usepackage{amsthm}
\usepackage{makecell}
\usepackage{float}

\usepackage[capitalize,noabbrev]{cleveref}

\theoremstyle{plain}

\theoremstyle{definition}

\theoremstyle{remark}

\usepackage[textsize=tiny]{todonotes}

\icmltitlerunning{Dividing and Conquering a BlackBox to a Mixture of Interpretable Models: Route, Interpret, Repeat}

\newcommand*\mystrut[1]{\vrule width0pt height0pt depth#1\relax}

\newcommand\ie {{\it i.e., }}
\newcommand\st {{\it s.t., }}
\newcommand\eg {{\it e.g., }}
\newcommand\etc{{\it etc.}}

\begin{document}

\twocolumn[
\icmltitle{Dividing and Conquering a BlackBox to a Mixture of Interpretable Models: Route, Interpret, Repeat}




\begin{icmlauthorlist}
\icmlauthor{Shantanu Ghosh}{bu}
\icmlauthor{Ke Yu}{pitt}
\icmlauthor{Forough Arabshahi}{meta}
\icmlauthor{Kayhan Batmanghelich}{bu}
\end{icmlauthorlist}

\icmlaffiliation{bu}{Department of Electrical and Computer Engineering, Boston University, MA, USA}
\icmlaffiliation{pitt}{Intelligent Systems Program, University of Pittsburgh, PA, USA}
\icmlaffiliation{meta}{MetaAI, MenloPark, CA, USA}

\icmlcorrespondingauthor{Shantanu Ghosh}{shawn24@bu.edu}

\icmlkeywords{Machine Learning, ICML}

\vskip 0.3in
]



\printAffiliationsAndNotice{} 

\begin{abstract}


ML model design either starts with an interpretable model or a Blackbox and explains it post hoc. Blackbox models are flexible but difficult to explain, while interpretable models are inherently explainable. Yet, interpretable models require extensive ML knowledge and tend to be less flexible and underperforming than their Blackbox variants. This paper aims to blur the distinction between a post hoc explanation of a Blackbox and constructing interpretable models. 
Beginning with a Blackbox, we iteratively \emph{carve out} a mixture of interpretable experts (MoIE) and a \emph{residual network}. Each interpretable model specializes in a subset of samples and explains them using First Order Logic (FOL), providing basic reasoning on concepts from the Blackbox. We route the remaining samples through a flexible residual. We repeat the method on the residual network until all the interpretable models explain the desired proportion of data. Our extensive experiments show that our \emph{route, interpret, and repeat} approach (1) identifies a diverse set of instance-specific concepts with high concept completeness via MoIE without compromising in performance, (2) identifies the relatively ``harder'' samples to explain via residuals, (3) outperforms the interpretable by-design models by significant margins during test-time interventions, and (4) fixes the shortcut learned by the original Blackbox. The code for MoIE is
publicly available at: \url{https://github.com/batmanlab/ICML-2023-Route-interpret-repeat}.
\end{abstract}

\section{Introduction}

Model explainability is essential in high-stakes applications of AI, \eg healthcare. While Blackbox models (\eg Deep Learning) offer flexibility and modular design, post hoc explanation is prone to confirmation bias~\cite{wan2022explainability}, lack of fidelity to the original model~\cite{adebayo2018sanity}, and insufficient mechanistic explanation of the decision-making process~\cite{rudin2019stop}. Interpretable-by-design models overcome those issues but tend to be less flexible than Blackbox models and demand substantial expertise to design. Using a post hoc explanation or adopting an inherently interpretable model is a mutually exclusive decision to be made at the initial phase of AI model design. This paper blurs the line on that dichotomous model design.

The literature on post hoc explanations is extensive. This includes model attributions (~\cite{simonyan2013deep, selvaraju2017grad}), counterfactual approaches ~\cite{abid2021meaningfully, singla2019explanation}, and distillation methods~\cite{alharbi2021learning, cheng2020explaining}. Those methods either identify key input features that contribute the most to the network's output~\cite{shrikumar2016not}, generate input perturbation to flip the network's output~\cite{samek2016evaluating, montavon2018methods}, or estimate simpler functions to approximate the network output locally. Post hoc methods preserve the flexibility and performance of the Blackbox but suffer from a lack of fidelity and mechanistic explanation of the network output~\cite{rudin2019stop}. Without a mechanistic explanation, recourse to a model's undesirable behavior is unclear. Interpretable models are alternative designs to the Blackbox without many such drawbacks. For example, modern interpretable methods highlight human understandable \emph{concepts} that contribute to the downstream prediction.

Several families of interpretable models exist for a long time, such as the rule-based approach and generalized additive models~\cite{hastie1987generalized, letham2015interpretable, breiman1984classification}. They primarily focus on tabular data. Such models for high-dimensional data (\eg images) primarily rely on projecting to a lower dimensional human understandable \emph{concept} or \emph{symbolic} space~\cite{koh2020concept} and predicting the output with an interpretable classifier. Despite their utility, the current State-Of-The-Art (SOTA)
are limited in design; for example, they do not model the interaction between the concepts except for a few exceptions~\cite{ciravegna2021logic, barbiero2022entropy}, offering limited reasoning capabilities and robustness. Furthermore, if a portion of the samples does not fit the template design of the interpretable model, they do not offer any flexibility, compromising performance. 

\textbf{Our contributions}
We propose an interpretable method, aiming to achieve the best of both worlds: not sacrificing Blackbox performance similar to post hoc explainability while still providing actionable interpretation. We hypothesize that a Blackbox encodes several interpretable models, each applicable to a different portion of data. Thus, a single interpretable model may be insufficient to explain all samples. We construct a hybrid neuro-symbolic model by progressively \emph{carving out} a mixture of interpretable models and a \emph{residual network} from the given Blackbox. We coin the term \emph{expert} for each interpretable model, as they specialize over a subset of data. All the interpretable models are termed a ``Mixture of Interpretable Experts'' (MoIE). Our design identifies a subset of samples and \emph{routes} them through the interpretable models to explain the samples with FOL, providing basic reasoning on concepts from the Blackbox. The remaining samples are routed through a flexible residual network. 
On the residual network, we repeat the method until MoIE explains the desired proportion of data.
We quantify the sufficiency of the identified concepts to explain the Blackbox’s prediction using the concept completeness score~\cite{yeh2019concept}.
Using FOL for interpretable models offers recourse when undesirable behavior is detected in the model. We provide an example of fixing a shortcut learning by modifying the FOL. FOL can be used in human-model interaction (not explored in this paper). Our method is the divide-and-conquer approach, where the instances covered by the residual network need progressively more complicated interpretable models. Such insight can be used to inspect the data and the model further. Finally, our model allows \emph{unexplainable} category of data, which is currently not allowed in the interpretable models.  



\section{Method}
\label{sec:method}

\textbf{Notation:} 
Assume we have a dataset \{$\mathcal{X}$, $\mathcal{Y}$, $\mathcal{C}$\}, where $\mathcal{X}$, $\mathcal{Y}$, and $\mathcal{C}$ are the input images, class labels, and human interpretable attributes, respectively. $\displaystyle f^0:  \mathcal{X} \rightarrow \mathcal{Y}$, is our pre-trained initial Blackbox model. We assume that $\displaystyle f^0$ is a composition $\displaystyle h^0 \circ \Phi $, where $\displaystyle \Phi: \mathcal{X} \rightarrow \mathbb{R}^l $ is the  image embeddings and $\displaystyle h^0: \mathbb{R}^l \rightarrow \mathcal{Y}$ is a transformation from the embeddings, $\Phi$, to the class labels. We denote the learnable function $\displaystyle t: \mathbb{R}^l \rightarrow\mathcal{C}$, projecting the image embeddings to the concept space.
The concept space is the space spanned by the attributes $\mathcal{C}$. Thus, function $t$ outputs a scalar value representing a concept for each input image. 

\begin{figure}[t]
\begin{center}
\centerline{\includegraphics[width=\columnwidth]{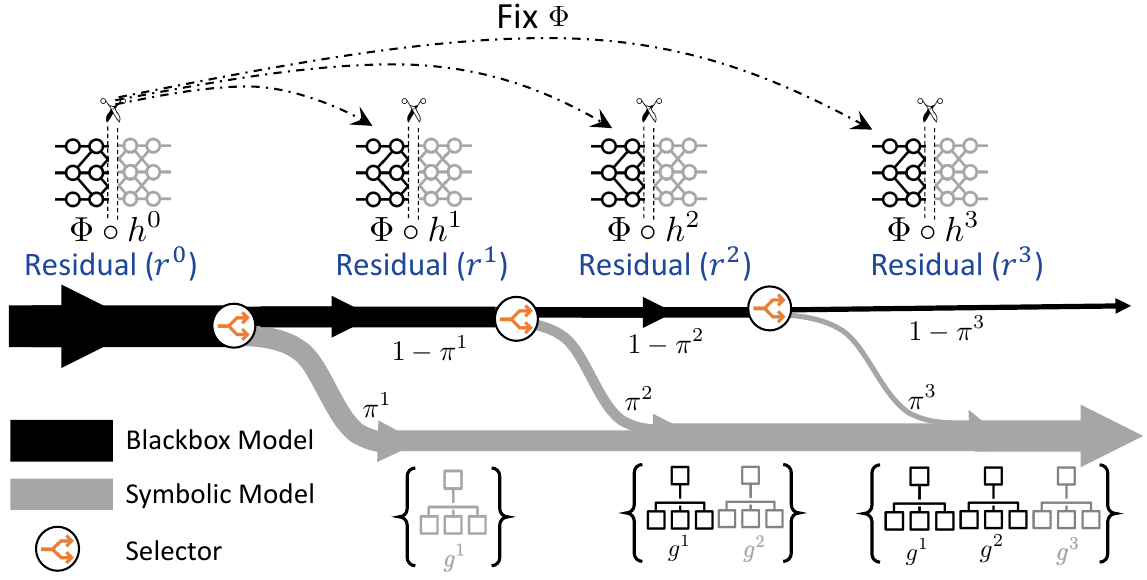}}
\caption{Schematic view of \emph{route, interpret} and \emph{repeat}. At iteration $k$, the selector \emph{routes} each sample either towards the interpretable model $g^k$ (to \emph{interpret}) with probability $\pi^k(.)$ or the residual $r^k = f^{k-1} - g^k$ with probability $1-\pi^k(.)$ (to 
\emph{repeat} in the further iterations). 
$f^{k-1}$ is the Blackbox of the $(k-1)^{th}$ iteration. $g^k$ generates FOL-based explanations for the samples it covers. 
Otherwise, the selector routes through the next step until it either goes through a subsequent interpretable model or reaches the last residual. 
Components in black and grey indicate the fixed and trainable modules in our model, respectively.
}
\label{fig:Schematic} 
\end{center}
\vskip -0.2in
\end{figure}

\textbf{Method Overview:} 
\cref{fig:Schematic} summarizes our approach. We iteratively carve out an interpretable model from the given Blackbox. Each iteration yields an interpretable model (the downward grey paths in~\cref{fig:Schematic}) and a residual (the straightforward black paths in~\cref{fig:Schematic}).
We start with the initial Blackbox $f^0$.
At iteration $k$, we distill the Blackbox from the previous iteration $f^{k-1}$ into a neuro-symbolic interpretable model, $\displaystyle g^{k}: \mathcal{C} \rightarrow \mathcal{Y}$. Our $g$ is flexible enough to be any interpretable models~\cite{yuksekgonul2022post, koh2020concept, barbiero2022entropy}. The \emph{residual} $r^k =f^{k-1} - g^k$ emphasizes the portion of $f^{k-1}$ that $g^k$cannot explain. We then approximate $r^k$ with $f^{k} = h^k \circ \Phi$. $f^k$ will be the Blackbox for the subsequent iteration and be explained by the respective interpretable model. 
A learnable gating mechanism, denoted by $\pi^k : \mathcal{C} \rightarrow \{0,1\}$ (shown as the \emph{selector} in~\cref{fig:Schematic}) routes an input sample towards either $g^k$ or $r^k$.
The thickness of the lines in~\cref{fig:Schematic} represents the samples covered by the interpretable models (grey line) and the residuals (black line). 
With every iteration, the cumulative coverage of the interpretable models increases, but the residual decreases. We name our method \emph{route, interpret} and \emph{repeat}.

\subsection{Neuro-Symbolic Knowledge Distillation}
Knowledge distillation in our method involves 3 parts: (1) a series of trainable selectors, \emph{routing} each sample through the interpretable models and the residual networks, (2) a sequence of learnable neuro-symbolic interpretable models, each providing FOL explanations to \emph {interpret} the Blackbox, and (3) \emph{repeating} with Residuals for the samples that cannot be explained with their interpretable counterparts. 
We detail each component below.

\subsubsection{The selector function}
As the first step of our method, the selector $\pi^k$ \emph{routes} the $j^{th}$ sample through the interpretable model $g^k$ or residual $r^k$ with probability $\displaystyle \pi^k(\boldsymbol{c_j})$ and $\displaystyle 1 - \pi^k(\boldsymbol{c_j})$ respectively, where $k$ $\in [0,K]$, with $K$ being the number of iterations.
We define the empirical coverage of the $\displaystyle k^{th}$ iteration as $\vspace{-0.08pt} \zeta(\pi^k) = \frac{1}{m}\sum_{j = 1} ^ m \pi^k(\boldsymbol{c_j}) \vspace{-0.081pt}$, the empirical mean of the samples selected by the selector for the associated interpretable model $\displaystyle g^k$, with $\displaystyle m$ being the total number of samples in the training set. Thus, the entire selective risk is:

\begin{equation}
\label{equ: emp_risk}
\mathcal{R}^k(\displaystyle \pi^k, \displaystyle g^k) = \frac{\frac{1}{m}\sum_{j=1}^m\mathcal{L}_{(g^k, \pi^k)}^k\big(\boldsymbol{x_j}, \boldsymbol{c_j}\big)}{\zeta(\pi^k)} ,
\end{equation}

where $\mathcal{L}_{(g^k, \pi^k)}^k$ is the optimization loss used to learn $\displaystyle g^k$ and $\displaystyle \pi^k$ together, discussed in~\cref{ns-optimization}. For a given coverage of $\displaystyle \tau^k \in (0, 1]$, we solve the following optimization problem:

\vskip -7.5pt
\begin{align}
\label{equ: optimization_g}
\theta_{s^k}^*, \theta_{g^k}^* = & \operatorname*{arg\,min}_{\theta_{s^k}, \theta_{g^k}} \mathcal{R}^k\Big(\pi^k(.; \theta_{s^k}), \displaystyle g^k(.; \theta_{g^k}) \Big) \nonumber \\ 
&\text{s.t.} ~~~ \zeta\big(\pi^k(.; \theta_{s^k})\big) \geq \tau^k,
\end{align}
\vskip 2pt

where $\theta_{s^k}^*, \theta_{g^k}^*$ are the optimal parameters at iteration $k$ for the selector $\pi^k$ and the interpretable model $g^k$ respectively. In this work, $\pi$s' of different iterations are neural networks with sigmoid activation. At inference time, the selector routes the $j^{th}$ sample with concept vector $\boldsymbol{c_j}$ to $\displaystyle g^k$ if and only if $\pi^k(\boldsymbol{c}_j)\geq 0.5$ for $k \in [0,K]$.

\subsubsection{Neuro-Symbolic interpretable models}
\label{ns-optimization}
In this stage, we design interpretable model $\displaystyle g^k$ of $k^{th}$ iteration  to \emph{interpret} the Blackbox $\displaystyle f^{k - 1}$ from the previous $(k-1)^{th}$ iteration by optimizing the following loss function:
\begin{equation}
\label{equ: g_k}
\resizebox{0.47\textwidth}{!}{$
\mathcal{L}_{(g^k, \pi^k)}^k\big(\boldsymbol{x_j}, \boldsymbol{c_j}\big) = \underbrace{\mystrut{2.6ex}\ell\Big(f^{k - 1}(\boldsymbol{x_j}), g^k(\boldsymbol{c_j})\Big)\pi^k(c_j) }_{\substack{\text{trainable component} \\ \text{for current iteration $k$}}}\underbrace{\prod_{i=1} ^{k - 1}\big(1 - \pi^i(\boldsymbol{c_j})\big)}_{\substack{\text{fixed component trained} \\ \text{in the previous iterations}}},
$}
\end{equation}

where the term $\pi^k(\boldsymbol{c_j})\prod_{i=1} ^{k - 1}\big(1 - \pi^i(\boldsymbol{c_j}) \big)$ denotes the probability of $j^{th}$ sample being routed through the interpretable model $g^k$. It is the probability of the sample going through the residuals for all the previous iterations from $1$ through $k-1$ (\ie $\prod_{i=1} ^{k - 1}\big(1 - \pi^i(\boldsymbol{c_j}) \big)$\big) times the probability of going through the interpretable model at iteration $k$ \big(\ie $\pi^k(\boldsymbol{c_j})$\big). 
Refer to~\cref{fig:Schematic} for an illustration. We learn $\pi^1, \dots \pi^{k - 1}$ in the prior iterations and are not trainable at iteration $k$. As each interpretable model $g^k$ specializes in explaining a specific subset of samples (denoted by coverage $\tau$), we refer to it as an \emph{expert}. We use SelectiveNet's ~\cite{geifman2019selectivenet}  optimization method to optimize~\cref{equ: optimization_g} since selectors need a rejection mechanism to route samples through residuals.~\cref{app:loss} details the optimization procedure in~\cref{equ: g_k}. We refer to the interpretable experts of all the iterations as a ``Mixture of Interpretable Experts'' (MoIE) cumulatively after training. Furthermore, we utilize E-LEN, \ie a Logic Explainable Network~\cite{ciravegna2023logic} implemented with an Entropy Layer as first layer~\cite{barbiero2022entropy} as the interpretable symbolic model $g$ to construct First Order Logic (FOL) explanations of a given prediction.

\subsubsection{The Residuals}
The last step is to \emph{repeat} with the residual $r^k$, as $\displaystyle r^k(\boldsymbol{x_j},\boldsymbol{c_j}) = f^{k - 1}(\boldsymbol{x_j}) - g^k(\boldsymbol{c_j})$.
We train $f^k = h^k\big(\Phi(.)\big)$ to approximate the residual $r^k$, creating a new Blackbox $f^k$ for the next iteration $(k+1$). This step is necessary to specialize $\displaystyle f^k$ over samples not covered by $g^k$. Optimizing the following loss function yields $\displaystyle f^k$ for the $\displaystyle k^{th}$ iteration:
\begin{equation}
\label{equ: residual}
\mathcal{L}_f^k(\boldsymbol{x_j}, \boldsymbol{c_j}) = \underbrace{\mystrut{2.6ex}\ell\big(r^k(\boldsymbol{x_j}, \boldsymbol{c_j}), f^k(\boldsymbol{x_j})\big)}_{\substack{\text{trainable component} \\ \text{for iteration $k$}}} \underbrace{\mystrut{2.6ex}\prod_{i=1} ^{k}\big(1 - \pi^i(\boldsymbol{c_j})\big)}_{\substack{\text{non-trainable component} \\ \text{for iteration $k$}}} 
\end{equation}

Notice that we fix the embedding $\displaystyle \Phi(.)$ for all the iterations. Due to computational overhead, we only finetune the last few layers of the Blackbox ($h^k$) to train $f^k$.
At the final iteration $K$, our method produces a MoIE and a Residual, explaining the interpretable and uninterpretable components of the initial Blackbox $f^0$, respectively.~\cref{app:algo} describes the training procedure of our model, the extraction of FOL, and the architecture of our model at inference.

\textbf{Selecting number of iterations $K$:} We follow two principles to select the number of iterations $K$ as a stopping criterion: 1) Each expert should have enough data to be trained reliably (
coverage $\zeta^k$). If an expert covers insufficient samples, we stop the process. 2) If the final residual ($r^K$) underperforms a threshold, it is not reliable to distill from the Blackbox. We stop the procedure to ensure that overall accuracy is maintained.

\begin{table}[t]
\caption{Datasets and Blackboxes.}
\fontsize{5.2pt}{0.30cm}\selectfont
\label{tab:dataset}
\vskip 0.1in
\begin{center}
\begin{tabular}{lcc}
\toprule
DATASET & BLACKBOX & \# EXPERTS  \\
\midrule
CUB-200~\cite{wah2011caltech}  & RESNET101~\cite{he2016deep}& 6 \\
CUB-200~\cite{wah2011caltech}  & VIT~\cite{wang2021feature}&  6 \\
AWA2~\cite{xian2018zero} & RESNET101~\cite{he2016deep} & 4\\
AWA2~\cite{xian2018zero} & VIT~\cite{wang2021feature} & 6\\
HAM1000~\cite{tschandl2018ham10000}    & INCEPTION~\cite{szegedy2015going}& 6\\
SIIM-ISIC~\cite{rotemberg2021patient}    & INCEPTION~\cite{szegedy2015going} & 6\\
EFFUSION IN MIMIC-CXR~\cite{12_johnsonmimic}    & DENSENET121~\cite{huang2017densely} & 3\\
\bottomrule
\end{tabular}
\end{center}
\vskip -0.1in
\end{table}

\begin{figure*}[t]
\vskip 0.1in
\begin{center}
\centerline{\includegraphics[width=\linewidth]{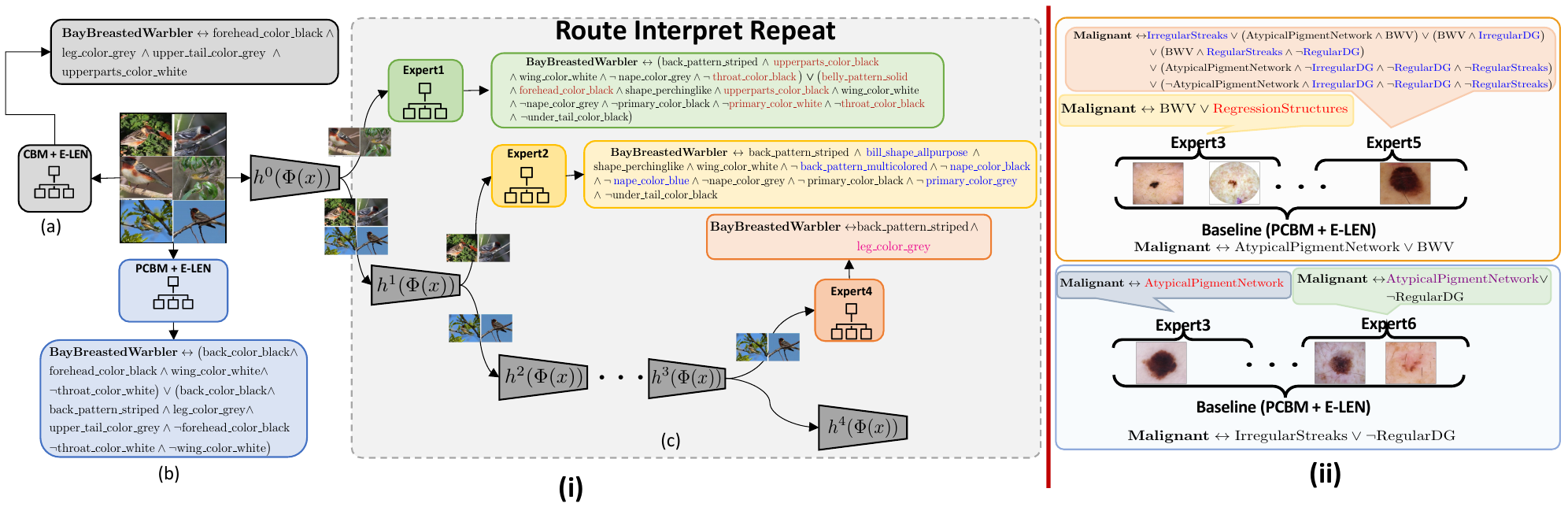}}
\caption{MoIE identifies diverse concepts for specific subsets of a class, unlike the generic ones by the baselines. \textbf{(i)} We construct the FOL explanations of the samples of, ``Bay breasted warbler'' in the CUB-200 dataset for VIT-based \textbf{(a)} CBM + E-LEN as an \emph{interpretable-by-design} baseline, \textbf{(b)} PCBM + E-LEN as a \emph{posthoc} baseline, \textbf{(c)} experts in MoIE at inference. We highlight the unique concepts for experts 1,2, and 3 in~\emph{red},~\emph{blue}, and~\emph{magenta}, respectively. \textbf{(ii)} Comparison of FOL explanations by MoIE with the PCBM + E-LEN baselines for HAM10000 (\textbf{top}) and ISIC (\textbf{down})  to classify Malignant lesion. We highlight unique concepts for experts 3, 5, and 6 in \emph{red}, \emph{blue}, and \emph{violet}, respectively. For brevity, we combine the local FOLs for each expert for the samples covered by them, shown in the figure.}
\label{fig:local_ex_cub}
\end{center}
\vskip -0.1in
\end{figure*}

\section{Related work}

\textbf{Post hoc explanations:} 
Post hoc explanations retain the flexibility and performance of the Blackbox. The post hoc explanation has many categories, including feature attribution~\cite{simonyan2013deep, smilkov2017smoothgrad, binder2016layer} and counterfactual approaches~\cite{singla2019explanation, abid2021meaningfully}. For example, feature attribution methods associate a measure of importance to features (e.g., pixels) that is proportional to the feature's contribution to BlackBox's predicted output. Many methods were proposed to estimate the importance measure, including gradient-based methods~\cite{selvaraju2017grad, sundararajan2017axiomatic}, game-theoretic approach~\cite{SHAP}. The post hoc approaches suffer from a lack of fidelity to input~\cite{adebayo2018sanity} and ambiguity in explanation due to a lack of correspondence to human-understandable concepts. Recently, Posthoc Concept Bottleneck models (PCBMs) ~\cite{yuksekgonul2022post} learn the concepts from a trained Blackbox embedding and use an interpretable classifier for classification. Also, they fit a residual in their hybrid variant (PCBM-h) to mimic the performance of the Blackbox. We will compare against the performance of the PCBMs method. Another major shortcoming is that, due to a lack of mechanistic explanation, post hoc explanations do not provide a recourse when an undesirable property of a Blackbox is identified. Interpretable-by-design provides a remedy to those issues~\cite{rudin2019stop}.

\textbf{Concept-based interpretable models:}
Our approach falls into the category of concept-based interpretable models. 
Such methods provide a mechanistically interpretable prediction that is a function of human-understandable concepts. The concepts are usually extracted from the activation of the middle layers of the Neural Network (bottleneck). Examples include Concept Bottleneck models (CBMs)~\cite{koh2020concept},  antehoc concept decoder~\cite{sarkar2021inducing}, and a high-dimensional Concept Embedding model (CEMs)~\cite{zarlenga2022concept} that uses high dimensional concept embeddings to allow extra supervised learning capacity and achieves SOTA performance in the interpretable-by-design class. Most concept-based interpretable models do not model the interaction between concepts and cannot be used for reasoning. An exception is E-LEN~\cite{barbiero2022entropy} which uses an entropy-based approach to derive explanations in terms of FOL using the concepts. The underlying assumption of those methods is that one interpretable function can explain the entire set of data, which can limit flexibility and consequently hurt the performance of the models. Our approach relaxes that assumption by allowing multiple interpretable functions and a residual. Each function is appropriate for a portion of the data, and a small portion of the data is allowed to be uninterpretable by the model (\ie residual). We will compare our method with CBMs, CEMs, and their E-LEN-enhanced variants.

\textbf{Application in fixing the shortcut learning:}
Shortcuts are spurious features that correlate with both input and the label on the training dataset but fail to generalize in more challenging real-world scenarios. Explainable AI (X-AI) aims to identify and fix such an undesirable property. Related work in X-AI includes LIME~\cite{ribeiro2016should}, utilized to detect spurious background as a shortcut to classify an animal. Recently interpretable model~\cite{rosenzweig2021patch}, involving local image patches, are used 
as a proxy to the Blackbox to identify shortcuts. However, both methods operate in pixel space, not concept space. Also, both approaches are post hoc and do not provide a way to eliminate the shortcut 
learning problem. Our MoIE discovers shortcuts using the high-level concepts in the FOL explanation of the Blackbox's prediction and eliminates them via metadata normalization (MDN)~\cite{lu2021metadata}.

\begin{figure*}[ht]
\vskip 0.2in
\begin{center}
\centerline{\includegraphics[width=\linewidth]{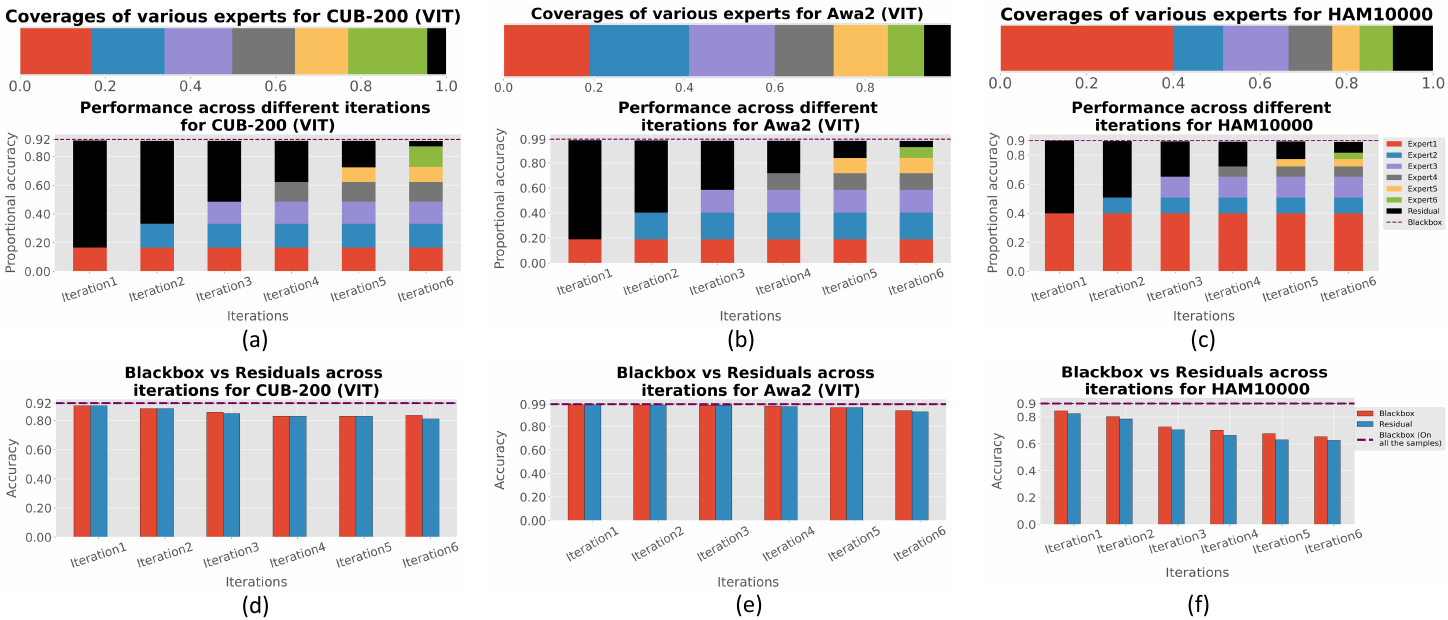}}
\caption{The performance of experts and residuals across iterations. 
\textbf{(a-c)} Coverage and proportional accuracy of the experts and residuals. 
\textbf{(d-f)} We route the samples covered by the residuals across iterations to the initial Blackbox $f^0$ and compare the accuracy of $f^0$ (red bar) with the residual (blue bar). Figures \textbf{d-f} show the progressive decline in performance of the residuals across iterations as they cover the samples in the increasing order of ``hardness''. We observe the similar abysmal performance of the initial blackbox $f^0$ for these samples.
}
\label{fig:expert_performance_cv_vit}
\end{center}
\vskip -0.2in
\end{figure*}

\section{Experiments}

We perform experiments on a variety of vision and medical imaging datasets to show that 1) MoIE captures a diverse set of concepts, 2) the performance of the residuals degrades over successive iterations as they cover ``harder'' instances, 3) MoIE does not compromise the performance of the Blackbox, 4) MoIE achieves superior performances during test time interventions, and 5) MoIE can fix the shortcuts using the Waterbirds dataset \cite{sagawa2019dro}. We repeat our method until MoIE covers at least 90\% of samples or the final residual's accuracy falls below 70\%. Furthermore, we only include concepts as
input to $g$ if their validation accuracy or auroc exceeds a certain threshold (in all of our experiments,
we fix 0.7 or 70\% as the threshold of validation auroc or accuracy). Refer to~\cref{tab:dataset} for the datasets and Blackboxes experimented with. For convolution based Blackboxes (ResNets, Densenet121 and Inception), we flatten the feature maps from the last convolutional block to extract the concepts. For VITs, we use the image embeddings from the transformer encoder to perform the same. We use SIIM-ISIC as a real-world transfer learning setting, with the Blackbox trained on HAM10000 and evaluated on a subset of the SIIM-ISIC Melanoma Classification dataset~\cite{yuksekgonul2022post}.~\cref{app:dataset} and~\cref{app:g} expand on the datasets and hyperparameters.

\begin{figure*}[ht]
\vskip 0.2in
\begin{center}
\centerline{\includegraphics[width=\linewidth]{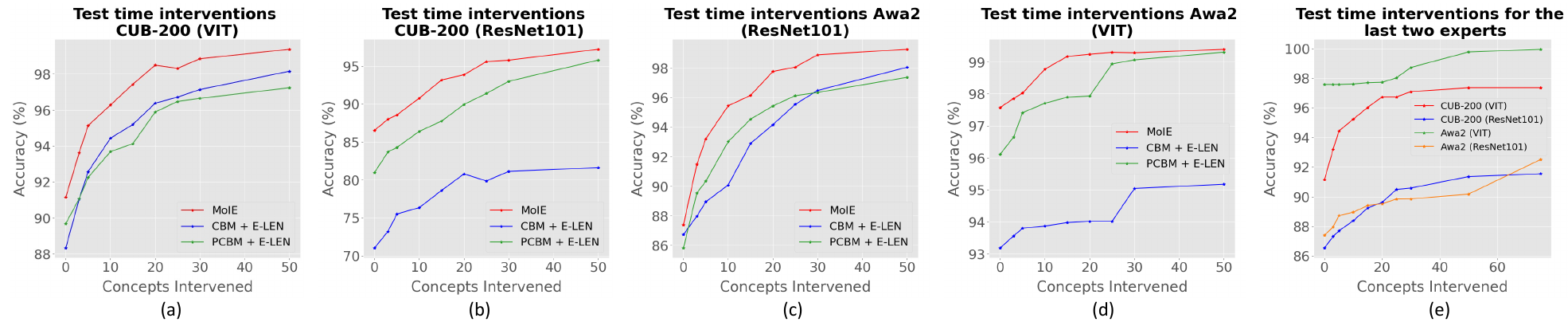}}
\caption{Across architectures test time interventions of concepts  on all the samples \textbf{(a-d)}, on the ``hard'' samples \textbf{(e)}, covered by only the last two experts of MoIE. 
}
\label{fig:tti}
\end{center}
\vskip -0.2in
\end{figure*}


\begin{figure}[h]
\begin{center}
\centerline{\includegraphics[width=\columnwidth]{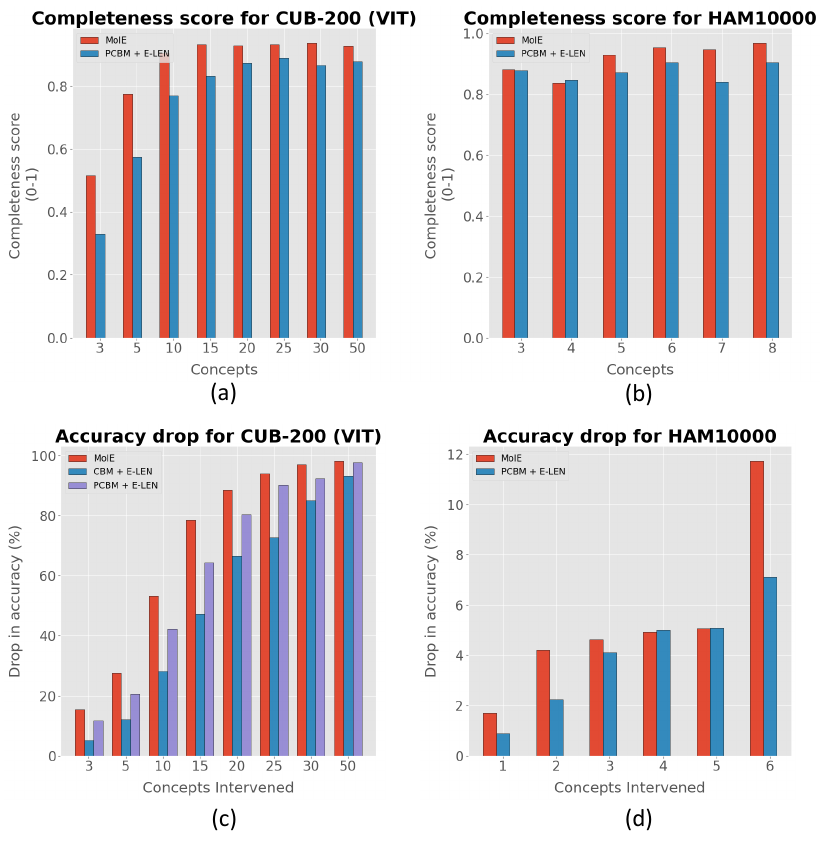}}
\caption{Quantitative validation of the extracted concepts. \textbf{(a-b)} 
Completeness scores of the models for a varying number of top concepts. (\textbf{c-d)} Drop in accuracy compared to the original model after zeroing out the top significant concepts iteratively. The highest drop for MoIE indicates that MoIE selects more instance-specific concepts than generic ones by the baselines. 
}
\label{fig:valid_concepts}
\end{center}
\vskip -0.2in
\end{figure}

\textbf{Baselines:}
We compare our methods to two concept-based baselines -- 1) interpretable-by-design and 2) posthoc. They consist of two parts: a) a concept predictor $\Phi: \mathcal{X} \rightarrow \mathcal{C}$, predicting concepts from images; and b) a label predictor $g: \mathcal{C} \rightarrow \mathcal{Y}$, predicting labels from the concepts. The end-to-end CEMs and sequential CBMs serve as interpretable-by-design baselines. Similarly, PCBM and PCBM-h serve as post hoc baselines. Convolution-based $\Phi$ includes all layers till the last convolution block. VIT-based $\Phi$ consists of the transformer encoder block. The standard CBM and PCBM models do not show how the concepts are composed to make the label prediction. So, we create CBM + E-LEN, PCBM + E-LEN and PCBM-h + E-LEN by using the identical $g$ of MOIE (shown in~\cref{app:g}), as a replacement for the standard classifiers of CBM and 
PCBM. We train the $\Phi$ and $g$ in these new baselines to sequentially generate FOLs~\cite{barbiero2022entropy}. Due to the unavailability of concept annotations, we extract the concepts from the Derm7pt dataset~\cite{kawahara2018seven} using the pre-trained embeddings of the Blackbox~\cite{yuksekgonul2022post} for HAM10000. Thus, we do not have interpretable-by-design baselines for HAM10000 and ISIC.

\subsection{Results}

\subsubsection{Expert driven explanations by MoIE }

First, we show that MoIE captures a rich set of diverse instance-specific concepts qualitatively. Next, we show quantitatively that MoIE-identified concepts are faithful to Blackbox's final prediction using the metric ``completeness score'' and zeroing out relevant concepts.

\textbf{Heterogenity of Explanations:}
At each iteration of MoIE, the blackbox \big($h^k(\Phi(.)$\big) splits into an interpretable expert ($g^k$) and a residual ($r^k$).~\cref{fig:local_ex_cub}i shows this mechanism for VIT-based MoIE and compares the FOLs with CBM + E-LEN and PCBM + E-LEN baselines to classify ``Bay Breasted Warbler'' of CUB-200.
The experts of different iterations specialize in specific instances of ``Bay Breasted Warbler''. Thus, each expert's FOL comprises its instance-specific concepts of the same class. For example, the concept, \emph{leg\_color\_grey} is unique to expert4, but \emph{belly\_pattern\_solid} and \emph{back\_pattern\_multicolored} are unique to experts 1 and 2, respectively, to classify the instances of ``Bay Breasted Warbler'' in the ~\cref{fig:local_ex_cub}(i)-c. 
Unlike MoIE, the baselines employ a single interpretable model $g$, resulting in a generic FOL with identical concepts for all the samples of ``Bay Breasted Warbler'' (\cref{fig:local_ex_cub}i(a-b)). Thus the baselines fail to capture the heterogeneity of explanations. Due to space constraint, we combine the local FOLs of different samples. For additional results of CUB-200, refer to~\cref{app:local_cub}.

\cref{fig:local_ex_cub}ii shows such diverse local instance-specific explanations for HAM10000 (\emph{top}) and ISIC (\emph{bottom}). In~\cref{fig:local_ex_cub}ii-(top), the baseline-FOL consists of concepts such as \emph{AtypicalPigmentNetwork} and \emph{BlueWhitishVeil (BWV)} to classify ``Malignancy'' for all the instances for HAM10000. However, expert~3 relies on \emph{RegressionStructures} along with \emph{BWV} to classify the same for the samples it covers while expert~5 utilizes several other concepts \eg \emph{IrregularStreaks}, \emph{Irregular dots and globules (IrregularDG)} \etc \text{ }Due to space constraints,~\cref{app:local_awa2} reports similar results for the Awa2 dataset. Also, VIT-based experts compose less concepts per sample than the ResNet-based experts, shown in~\cref{app:comparison_arch}.

\begin{table*}[t]
\caption{MoIE does not hurt the performance of the original Blackbox using a held-out test set. We 
provide the mean and standard errors of AUROC and accuracy for medical imaging (\eg HAM10000, ISIC, and Effusion) and vision (\eg CUB-200 and Awa2) datasets, respectively, over 5 random seeds. For MoIE, we also report the percentage of test set samples covered by all experts as ``coverage''. Here, MoIE + Residual represents the experts with the final residual. Following the setting~\cite{zarlenga2022concept}, we only report the performance of the convolutional CEM, leaving the construction of VIT-based CEM as a future work. Recall that interpretable-by-design models can not be constructed for HAM10000 and ISIC as they have no concept annotation; we learn the concepts from the Derm7pt dataset. For all the datasets, MoIE covers a significant portion of data (at least 90\%) cumulatively. We boldface our results. 
}
\fontsize{4.5pt}{0.28cm}\selectfont
\label{tab:performance}
\begin{center}
\begin{tabular}{p{25.2em} p{9em} p{9em} p{9em} p{9em} p{9em} p{9em} p{9em}}
\toprule 
        \textbf{MODEL} & \multicolumn{7}{c}{\textbf{DATASET}} \\
       & CUB-200 (RESNET101) & CUB-200 (VIT) & AWA2 (RESNET101) & AWA2 (VIT) & HAM10000 & SIIM-ISIC & EFFUSION  \\
\midrule 
    BLACKBOX & 0.88 & 0.92 & 0.89 & 0.99 & 0.96 & 0.85 & 0.91\\
\midrule
    \textbf{INTERPRETABLE-BY-DESIGN} \\
    CEM~\cite{zarlenga2022concept} & 0.77 $\pm$ 0.002 & - & 0.88 $\pm$ 0.005 & - & NA & NA & 0.76 $\pm$ 0.002\\
    CBM (Sequential)~\cite{koh2020concept} & 0.65 $\pm$ 0.003 & 0.86 $\pm$ 0.002 & 0.88 $\pm$ 0.003 & 0.94 $\pm$ 0.002  & NA & NA  
    & 0.79 $\pm$ 0.005 \\ 
    CBM + E-LEN~\cite{koh2020concept, barbiero2022entropy} & 0.71 $\pm$ 0.003 & 0.88 $\pm$ 0.002 & 0.86 $\pm$ 0.003 & 0.93 $\pm$ 0.002 & NA & NA & 
    0.79 $\pm$ 0.002  \\
\midrule
     \textbf{POSTHOC} \\
     PCBM~\cite{yuksekgonul2022post} & 0.76 $\pm$ 0.001  & 0.85 $\pm$ 0.002 & 0.82 $\pm$ 0.002 & 0.94 $\pm$ 0.001 &
     0.93 $\pm$	0.001 & 0.71 $\pm$	0.012 & 0.81 $\pm$	0.017\\
     PCBM-h~\cite{yuksekgonul2022post} & 0.85 $\pm$ 0.001  & 0.91 $\pm$ 0.001 & 0.87 $\pm$ 0.002 & 0.98 $\pm$ 0.001 &
     0.95 $\pm$	0.001 & 0.79 $\pm$	0.056 & 0.87 $\pm$	0.072\\
     PCBM + E-LEN~\cite{yuksekgonul2022post, barbiero2022entropy} &  0.80 $\pm$ 0.003 & 0.89 $\pm$ 0.002 & 0.85 $\pm$ 0.002 & 0.96 $\pm$ 0.001 & 
     0.94 $\pm$	0.021 &  0.73 $\pm$	0.011 & 0.81 $\pm$	0.014\\
     PCBM-h + E-LEN~\cite{yuksekgonul2022post, barbiero2022entropy} &  0.85 $\pm$ 0.003 & 0.91 $\pm$ 0.002 & 0.88 $\pm$ 0.002 & 0.98 $\pm$ 0.002 & 
     0.95 $\pm$	0.032 &  0.82 $\pm$	0.056 & 0.87 $\pm$	0.032\\
\midrule
     \textbf{OURS} \\
     MoIE (COVERAGE) &\textbf{0.86 $\pm$ 0.001 (0.9)} &\textbf{0.91 $\pm$ 0.001 (0.95)} &
     \textbf{0.87 $\pm$ 0.002 (0.91)} & \textbf{0.97 $\pm$ 0.004 (0.94)} & \textbf{0.95 $\pm$	0.001 (0.9)}
     & \textbf{0.84 $\pm$ 0.001 (0.94)} & \textbf{0.87 $\pm$	0.001 (0.98)}\\
     MoIE + RESIDUAL & \textbf{0.84 $\pm$ 0.001} & \textbf{0.90 $\pm$ 0.001} & \textbf{0.86 $\pm$ 0.002} & \textbf{0.94 $\pm$ 0.004}
     & \textbf{0.92 $\pm$	0.00} & \textbf{0.82 $\pm$	0.01} & \textbf{0.86 $\pm$	0.00} \\
\bottomrule
\end{tabular}
\end{center}
\end{table*}

\textbf{MoIE-identified concepts attain higher completeness scores.} 
\cref{fig:valid_concepts}(a-b) shows the completeness scores~\cite{yeh2019concept} for varying number of concepts.
 Completeness score is a post hoc measure, signifying the identified concepts as ``sufficient statistic'' of the predictive capability of the Blackbox. Recall that $g$ utilizes E-LEN~\cite{barbiero2022entropy}, associating each concept with an attention weight after training. A concept with high attention weight implies its high predictive significance.
 Iteratively, we select the top relevant concepts based on their attention weights and compute the completeness scores for the top concepts for MoIE and the PCBM + E-LEN baseline in \cref{fig:valid_concepts}(a-b) (~\cref{app:completeness} for details).
 For example, MoIE achieves a completeness score of 0.9 compared to 0.75 of the baseline($\sim 20\%\uparrow$) for the 10 most significant concepts for the CUB-200 dataset with VIT as Blackbox.
 
\textbf{MoIE identifies more meaningful instance-specific concepts.} 
\cref{fig:valid_concepts}(c-d) reports the drop in accuracy by zeroing out the significant concepts.
Any interpretable model ($g$) supports concept-intervention~\cite{koh2020concept}. 
After identifying the top concepts from $g$ using the attention weights, as in the last section, we set these concepts' values to zero, compute the model's accuracy drop, and plot in~\cref{fig:valid_concepts}(c-d). When zeroing out the top 10 essential concepts for VIT-based CUB-200 models, MoIE records a drop of 53\% compared to 28\% and 42\% for the CBM + E-LEN and PCBM + E-LEN baselines, respectively, showing the faithfulness of the identified concepts to the prediction.

In both of the last experiments, MoIE outperforms the baselines as the baselines mark the same concepts as significant for all samples of each class. However,
MoIE leverages various experts specializing in different subsets of samples of different classes. 
For results of MIMIC-CXR and Awa2, refer to~\cref{app:mimic_cxr} and~\cref{app:awa2} respectively.

\subsubsection{Identification of harder samples by successive residuals}
\label{Sec:residual}
\cref{fig:expert_performance_cv_vit} (a-c) display the proportional accuracy of the experts and the residuals of our method per iteration. The proportional accuracy of each model (experts and/or residuals) is defined as the accuracy of that model times its coverage. Recall that the model's coverage is the empirical mean of the samples selected by the selector. 
\cref{fig:expert_performance_cv_vit}a show that the experts and residual cumulatively achieve an accuracy $\sim$ 0.92 for the CUB-200 dataset in iteration 1, with more contribution from the residual (black bar) than the expert1 (blue bar). Later iterations cumulatively increase and worsen the performance of the experts and corresponding residuals, respectively. The final iteration carves out the entire interpretable portion from the Blackbox $f^0$ via all the experts, resulting in their more significant contribution to the cumulative performance. The residual of the last iteration covers the ``hardest'' samples, achieving low accuracy. Tracing these samples back to the original Blackbox $f^0$, it also classifies these samples poorly (\cref{fig:expert_performance_cv_vit}{(d-f)}).
As shown in the coverage plot, this experiment reinforces~\cref{fig:Schematic}, where the flow through the experts gradually becomes thicker compared to the narrower flow of the residual with every iteration. Refer to~\cref{fig:expert_performance_cv_resnet} in the~\cref{app:resnet_cv} for the results of the ResNet-based MoIEs.

\subsubsection{Quantitative analysis of MoIE with the blackbox and baseline}

\textbf{Comparing with the interpretable-by-design baselines:}~\cref{tab:performance} shows that MoIE achieves comparable performance to the Blackbox. Recall that ``MoIE'' refers to the mixture of all interpretable experts ($g$) only excluding any residuals.
MoIE outperforms the interpretable-by-design baselines for all the datasets except Awa2. Since Awa2 is designed for zero-shot learning, its rich concept annotation makes it appropriate for interpretable-by-design models. In general, VIT-derived MoIEs perform better than their ResNet-based variants.

\textbf{Comparing with the PCBMs:}~\cref{tab:performance} shows that interpretable MoIE outperforms the interpretable posthoc baselines -- PCBM and PCBM + E-LEN for all the datasets, especially by a significant margin for CUB-200 and ISIC.
 We also report ``MoIE + Residual'' as the mixture of interpretable experts plus the final residual to compare with the residualized PCBM, \ie PCBM-h.~\cref{tab:performance} shows that PCBM-h performs slightly better than MoIE + Residual. Note that PCBM-h learns the residual by fitting the complete dataset to fix the interpretable PCBM's mistakes to replicate the performance of the Blackbox, resulting in better performance for PCBM-h than PCBM. However, we assume the Blackbox to be a combination of interpretable and uninterpretable components. So, we train the experts and the final residual to cover the interpretable and uninterpretable portions of the Blackbox respectively. In each iteration, our method learns the residuals to focus on the samples, which are not covered by the respective interpretable experts. Therefore, residuals are not designed to fix the mistakes made by the experts. In doing so, the final residual in MoIE + Residual covers the ``hardest'' examples, lowering its overall performance compared to MoIE. 

\subsubsection{Test time interventions}

\cref{fig:tti}(a-d) shows effect of test time interventions.
Any concept-based models~\cite{koh2020concept, zarlenga2022concept} allow test time interventions for datasets with concept annotation (\eg CUB-200, Awa2). We identify the significant concepts via their attention scores in $g$, as during the computation of completeness scores, and set their values with the ground truths, considering the ground truth concepts as an oracle.
 As MoIE identifies a more diverse set of concepts by focusing on different subsets of classes, MoIE outperforms the baselines in terms of accuracy for such test time interventions. Instead of manually deciding the samples to intervene, it is generally preferred to intervene on the ``harder'' samples, making the process efficient. As per~\cref{Sec:residual}, experts of different iterations cover samples with increasing order of ``hardness''. To intervene efficiently, we perform identical test-time interventions with varying numbers of concepts for the ``harder'' samples covered by the final two experts and plot the accuracy in~\cref{fig:tti}(e). For the VIT-derived MoIE of CUB-200, intervening only on 20 concepts enhances the accuracy of MoIE from 91\% to 96\% ($\sim 6.1\% \uparrow$). We cannot perform the same for the baselines as they cannot directly estimate ``harder" samples. Also, \cref{fig:tti} shows a relatively higher gain for ResNet-based models in general.~\cref{app:tti_qual} demonstrates an example of test time intervention of concepts for relatively ``harder'' samples, identified by the last two experts of MoIE.

\subsubsection{Application in the removal of shortcuts}
\begin{figure}[ht]
\begin{center}
\centerline{\includegraphics[width=\columnwidth]{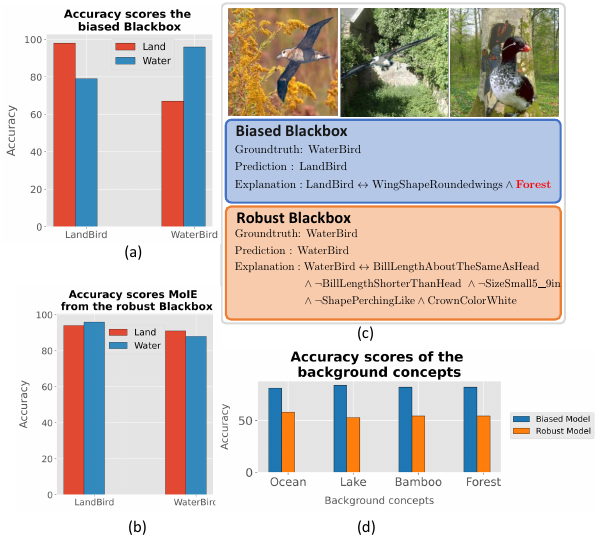}}
\caption{MoIE fixes shortcuts. \textbf{(a)} Performance of the biased Blackbox. 
\textbf{(b)} Performance of final MoIE extracted from the robust Blackbox after removing the shortcuts using MDN. 
\textbf{(c)} Examples of samples (\textbf{top-row}) and their explanations by the biased (\textbf{middle-row}) and robust Blackboxes (\textbf{bottom-row}). 
\textbf{(d)} Comparison of accuracies of the spurious concepts extracted from the biased vs. the robust Blackbox.
}
\label{fig:shortcut}
\end{center}
\vskip -0.2in
\end{figure}

First, we create the Waterbirds dataset as in~\cite{sagawa2019dro}by using forest and bamboo as the spurious land concepts of the Places dataset for landbirds of the CUB-200 dataset. We do the same by using oceans and lakes as the spurious water concepts for waterbirds. We utilize ResNet50 as the Blackbox $f^0$ to identify each bird as a Waterbird or a Landbird. The Blackbox quickly latches on the spurious backgrounds to classify the birds. As a result, the black box's accuracy differs for land-based versus aquatic subsets of the bird species, as shown in~\cref{fig:shortcut}a. The Waterbird on the water is more accurate than on land (96\%  vs. 67\% in the red bar in the~\cref{fig:shortcut}a). The FOL from the biased Blackbox-derived MoIE captures the spurious concept \textit{forest} for a waterbird, misclassified as a landbird. Assuming the background concepts as metadata, we remove the background bias from the representation of the Blackbox using Metadata normalization (MDN) layers~\cite{lu2021metadata} between two successive layers of the convolutional backbone to fine-tune the biased Blackbox to make it more robust. Next, we train $t$, using the embedding $\Phi$ of the robust Blackbox, and compare the accuracy of the spurious concepts with the biased blackbox in~\cref{fig:shortcut}d. The validation accuracy of all the spurious concepts retrieved from the robust Blackbox falls well short of the predefined threshold 70\% compared to the biased Blackbox. Finally, we re-train the MoIE distilling from the new robust Blackbox.~\cref{fig:shortcut}b illustrates similar accuracies of MoIE for Waterbirds on water vs. Waterbirds on land (89\% - 91\%). The FOL from the robust Blackbox does not include any background concepts (~\ref{fig:shortcut}c, bottom row). Refer to~\ref{fig:spurious_flow} in~\cref{app:shortcut} for the flow diagram of this experiment.

\section{Discussion \& Conclusions}

This paper proposes a novel method to iteratively extract a mixture of interpretable models from a flexible Blackbox. The comprehensive experiments on various datasets demonstrate that our method 1) captures more meaningful instance-specific concepts with high completeness score than baselines without losing the performance of the Blackbox, 2) does not require explicit concept annotation, 3) identifies the ``harder'' samples using the residuals, 4) achieves significant performance gain than the baselines during test time interventions, 5) eliminate shortcuts effectively. In the future, we aim to apply our method to other modalities, such as text or video. Also, as in the prior work, MoIE-captured concepts may not reflect a causal effect. The assessment of causal concept effects necessitates estimating inter-concept interactions, which will be the subject of future research.

\section{Acknowledgement}
We would like to thank Mert Yuksekgonul of Stanford University for providing the code to construct the concept bank of Derm7pt for conducting the skin experiments. This work was partially supported by NIH Award Number 1R01HL141813-01 and the Pennsylvania Department of Health. We are grateful for the computational resources provided by Pittsburgh Super Computing grant number TG-ASC170024.

\bibliography{example_paper}

\begin{thebibliography}{54}
\providecommand{\natexlab}[1]{#1}
\providecommand{\url}[1]{\texttt{#1}}
\expandafter\ifx\csname urlstyle\endcsname\relax
  \providecommand{\doi}[1]{doi: #1}\else
  \providecommand{\doi}{doi: \begingroup \urlstyle{rm}\Url}\fi

\bibitem[Abid et~al.(2021)Abid, Yuksekgonul, and Zou]{abid2021meaningfully}
Abid, A., Yuksekgonul, M., and Zou, J.
\newblock Meaningfully explaining model mistakes using conceptual
  counterfactuals.
\newblock \emph{arXiv preprint arXiv:2106.12723}, 2021.

\bibitem[Adebayo et~al.(2018)Adebayo, Gilmer, Muelly, Goodfellow, Hardt, and
  Kim]{adebayo2018sanity}
Adebayo, J., Gilmer, J., Muelly, M., Goodfellow, I., Hardt, M., and Kim, B.
\newblock Sanity checks for saliency maps.
\newblock \emph{Advances in neural information processing systems}, 31, 2018.

\bibitem[Alharbi et~al.(2021)Alharbi, Vu, and Thai]{alharbi2021learning}
Alharbi, R., Vu, M.~N., and Thai, M.~T.
\newblock Learning interpretation with explainable knowledge distillation.
\newblock In \emph{2021 IEEE International Conference on Big Data (Big Data)},
  pp.\  705--714. IEEE, 2021.

\bibitem[Barbiero et~al.(2022)Barbiero, Ciravegna, Giannini, Li{\'o}, Gori, and
  Melacci]{barbiero2022entropy}
Barbiero, P., Ciravegna, G., Giannini, F., Li{\'o}, P., Gori, M., and Melacci,
  S.
\newblock Entropy-based logic explanations of neural networks.
\newblock In \emph{Proceedings of the AAAI Conference on Artificial
  Intelligence}, volume~36, pp.\  6046--6054, 2022.

\bibitem[Belle(2020)]{belle2020symbolic}
Belle, V.
\newblock Symbolic logic meets machine learning: A brief survey in infinite
  domains.
\newblock In \emph{International Conference on Scalable Uncertainty
  Management}, pp.\  3--16. Springer, 2020.

\bibitem[Besold et~al.(2017)Besold, Garcez, Bader, Bowman, Domingos, Hitzler,
  K{\"u}hnberger, Lamb, Lowd, Lima, et~al.]{besold2017neural}
Besold, T.~R., Garcez, A.~d., Bader, S., Bowman, H., Domingos, P., Hitzler, P.,
  K{\"u}hnberger, K.-U., Lamb, L.~C., Lowd, D., Lima, P. M.~V., et~al.
\newblock Neural-symbolic learning and reasoning: A survey and interpretation.
\newblock \emph{arXiv preprint arXiv:1711.03902}, 2017.

\bibitem[Binder et~al.(2016)Binder, Montavon, Lapuschkin, M{\"u}ller, and
  Samek]{binder2016layer}
Binder, A., Montavon, G., Lapuschkin, S., M{\"u}ller, K.-R., and Samek, W.
\newblock Layer-wise relevance propagation for neural networks with local
  renormalization layers.
\newblock In \emph{International Conference on Artificial Neural Networks},
  pp.\  63--71. Springer, 2016.

\bibitem[Breiman et~al.(1984)Breiman, Friedman, Stone, and
  Olshen]{breiman1984classification}
Breiman, L., Friedman, J., Stone, C., and Olshen, R.
\newblock Classification and regression trees (crc, boca raton, fl).
\newblock 1984.

\bibitem[Cheng et~al.(2020)Cheng, Rao, Chen, and Zhang]{cheng2020explaining}
Cheng, X., Rao, Z., Chen, Y., and Zhang, Q.
\newblock Explaining knowledge distillation by quantifying the knowledge.
\newblock In \emph{Proceedings of the IEEE/CVF conference on computer vision
  and pattern recognition}, pp.\  12925--12935, 2020.

\bibitem[Ciravegna et~al.(2021)Ciravegna, Barbiero, Giannini, Gori, Li{\'o},
  Maggini, and Melacci]{ciravegna2021logic}
Ciravegna, G., Barbiero, P., Giannini, F., Gori, M., Li{\'o}, P., Maggini, M.,
  and Melacci, S.
\newblock Logic explained networks.
\newblock \emph{arXiv preprint arXiv:2108.05149}, 2021.

\bibitem[Ciravegna et~al.(2023)Ciravegna, Barbiero, Giannini, Gori, Li{\'o},
  Maggini, and Melacci]{ciravegna2023logic}
Ciravegna, G., Barbiero, P., Giannini, F., Gori, M., Li{\'o}, P., Maggini, M.,
  and Melacci, S.
\newblock Logic explained networks.
\newblock \emph{Artificial Intelligence}, 314:\penalty0 103822, 2023.

\bibitem[Daneshjou et~al.(2021)Daneshjou, Vodrahalli, Liang, Novoa, Jenkins,
  Rotemberg, Ko, Swetter, Bailey, Gevaert, et~al.]{daneshjou2021disparities}
Daneshjou, R., Vodrahalli, K., Liang, W., Novoa, R.~A., Jenkins, M., Rotemberg,
  V., Ko, J., Swetter, S.~M., Bailey, E.~E., Gevaert, O., et~al.
\newblock Disparities in dermatology ai: Assessments using diverse clinical
  images.
\newblock \emph{arXiv preprint arXiv:2111.08006}, 2021.

\bibitem[Garcez et~al.(2015)Garcez, Besold, De~Raedt, F{\"o}ldiak, Hitzler,
  Icard, K{\"u}hnberger, Lamb, Miikkulainen, and Silver]{garcez2015neural}
Garcez, A.~d., Besold, T.~R., De~Raedt, L., F{\"o}ldiak, P., Hitzler, P.,
  Icard, T., K{\"u}hnberger, K.-U., Lamb, L.~C., Miikkulainen, R., and Silver,
  D.~L.
\newblock Neural-symbolic learning and reasoning: contributions and challenges.
\newblock In \emph{2015 AAAI Spring Symposium Series}, 2015.

\bibitem[Geifman \& El-Yaniv(2019)Geifman and
  El-Yaniv]{geifman2019selectivenet}
Geifman, Y. and El-Yaniv, R.
\newblock Selectivenet: A deep neural network with an integrated reject option.
\newblock In \emph{International conference on machine learning}, pp.\
  2151--2159. PMLR, 2019.

\bibitem[Hastie \& Tibshirani(1987)Hastie and
  Tibshirani]{hastie1987generalized}
Hastie, T. and Tibshirani, R.
\newblock Generalized additive models: some applications.
\newblock \emph{Journal of the American Statistical Association}, 82\penalty0
  (398):\penalty0 371--386, 1987.

\bibitem[Havasi et~al.(2022)Havasi, Parbhoo, and
  Doshi-Velez]{havasi2022addressing}
Havasi, M., Parbhoo, S., and Doshi-Velez, F.
\newblock Addressing leakage in concept bottleneck models.
\newblock In \emph{Advances in Neural Information Processing Systems}, 2022.

\bibitem[He et~al.(2016)He, Zhang, Ren, and Sun]{he2016deep}
He, K., Zhang, X., Ren, S., and Sun, J.
\newblock Deep residual learning for image recognition.
\newblock In \emph{Proceedings of the IEEE conference on computer vision and
  pattern recognition}, pp.\  770--778, 2016.

\bibitem[Hinton et~al.(2015)Hinton, Vinyals, Dean,
  et~al.]{hinton2015distilling}
Hinton, G., Vinyals, O., Dean, J., et~al.
\newblock Distilling the knowledge in a neural network.
\newblock \emph{arXiv preprint arXiv:1503.02531}, 2\penalty0 (7), 2015.

\bibitem[Huang et~al.(2017)Huang, Liu, Van Der~Maaten, and
  Weinberger]{huang2017densely}
Huang, G., Liu, Z., Van Der~Maaten, L., and Weinberger, K.~Q.
\newblock Densely connected convolutional networks.
\newblock In \emph{Proceedings of the IEEE conference on computer vision and
  pattern recognition}, pp.\  4700--4708, 2017.

\bibitem[Jain et~al.(2021)Jain, Agrawal, Saporta, Truong, Duong, Bui, Chambon,
  Zhang, Lungren, Ng, et~al.]{10_jain2021radgraph}
Jain, S., Agrawal, A., Saporta, A., Truong, S.~Q., Duong, D.~N., Bui, T.,
  Chambon, P., Zhang, Y., Lungren, M.~P., Ng, A.~Y., et~al.
\newblock Radgraph: Extracting clinical entities and relations from radiology
  reports.
\newblock \emph{arXiv preprint arXiv:2106.14463}, 2021.

\bibitem[Johnson et~al.()Johnson, Lungren, Peng, Lu, Mark, Berkowitz, and
  Horng]{12_johnsonmimic}
Johnson, A., Lungren, M., Peng, Y., Lu, Z., Mark, R., Berkowitz, S., and Horng,
  S.
\newblock Mimic-cxr-jpg-chest radiographs with structured labels.

\bibitem[Kawahara et~al.(2018)Kawahara, Daneshvar, Argenziano, and
  Hamarneh]{kawahara2018seven}
Kawahara, J., Daneshvar, S., Argenziano, G., and Hamarneh, G.
\newblock Seven-point checklist and skin lesion classification using multitask
  multimodal neural nets.
\newblock \emph{IEEE journal of biomedical and health informatics}, 23\penalty0
  (2):\penalty0 538--546, 2018.

\bibitem[Kim et~al.(2017)Kim, Wattenberg, Gilmer, Cai, Wexler, Viegas, and
  Sayres]{kim2017interpretability}
Kim, B., Wattenberg, M., Gilmer, J., Cai, C., Wexler, J., Viegas, F., and
  Sayres, R.
\newblock Interpretability beyond feature attribution: Quantitative testing
  with concept activation vectors (tcav).(2017).
\newblock \emph{arXiv preprint arXiv:1711.11279}, 2017.

\bibitem[Koh et~al.(2020)Koh, Nguyen, Tang, Mussmann, Pierson, Kim, and
  Liang]{koh2020concept}
Koh, P.~W., Nguyen, T., Tang, Y.~S., Mussmann, S., Pierson, E., Kim, B., and
  Liang, P.
\newblock Concept bottleneck models.
\newblock In \emph{International Conference on Machine Learning}, pp.\
  5338--5348. PMLR, 2020.

\bibitem[Letham et~al.(2015)Letham, Rudin, McCormick, and
  Madigan]{letham2015interpretable}
Letham, B., Rudin, C., McCormick, T.~H., and Madigan, D.
\newblock Interpretable classifiers using rules and bayesian analysis: Building
  a better stroke prediction model.
\newblock \emph{The Annals of Applied Statistics}, 9\penalty0 (3):\penalty0
  1350--1371, 2015.

\bibitem[Lu et~al.(2021)Lu, Zhao, Zhang, Pohl, Fei-Fei, Niebles, and
  Adeli]{lu2021metadata}
Lu, M., Zhao, Q., Zhang, J., Pohl, K.~M., Fei-Fei, L., Niebles, J.~C., and
  Adeli, E.
\newblock Metadata normalization.
\newblock In \emph{Proceedings of the IEEE/CVF Conference on Computer Vision
  and Pattern Recognition}, pp.\  10917--10927, 2021.

\bibitem[Lucieri et~al.(2020)Lucieri, Bajwa, Braun, Malik, Dengel, and
  Ahmed]{lucieri2020interpretability}
Lucieri, A., Bajwa, M.~N., Braun, S.~A., Malik, M.~I., Dengel, A., and Ahmed,
  S.
\newblock On interpretability of deep learning based skin lesion classifiers
  using concept activation vectors.
\newblock In \emph{2020 international joint conference on neural networks
  (IJCNN)}, pp.\  1--10. IEEE, 2020.

\bibitem[Lundberg \& Lee(2017)Lundberg and Lee]{SHAP}
Lundberg, S.~M. and Lee, S.-I.
\newblock A unified approach to interpreting model predictions.
\newblock In \emph{Proceedings of the 31st international conference on neural
  information processing systems}, pp.\  4768--4777, 2017.

\bibitem[Mendelson(2009)]{mendelson2009introduction}
Mendelson, E.
\newblock \emph{Introduction to mathematical logic}.
\newblock Chapman and Hall/CRC, 2009.

\bibitem[Montavon et~al.(2018)Montavon, Samek, and
  M{\"u}ller]{montavon2018methods}
Montavon, G., Samek, W., and M{\"u}ller, K.-R.
\newblock Methods for interpreting and understanding deep neural networks.
\newblock \emph{Digital signal processing}, 73:\penalty0 1--15, 2018.

\bibitem[Ribeiro et~al.(2016)Ribeiro, Singh, and Guestrin]{ribeiro2016should}
Ribeiro, M.~T., Singh, S., and Guestrin, C.
\newblock " why should i trust you?" explaining the predictions of any
  classifier.
\newblock In \emph{Proceedings of the 22nd ACM SIGKDD international conference
  on knowledge discovery and data mining}, pp.\  1135--1144, 2016.

\bibitem[Rosenzweig et~al.(2021)Rosenzweig, Sicking, Houben, Mock, and
  Akila]{rosenzweig2021patch}
Rosenzweig, J., Sicking, J., Houben, S., Mock, M., and Akila, M.
\newblock Patch shortcuts: Interpretable proxy models efficiently find
  black-box vulnerabilities.
\newblock In \emph{Proceedings of the IEEE/CVF Conference on Computer Vision
  and Pattern Recognition}, pp.\  56--65, 2021.

\bibitem[Rotemberg et~al.(2021)Rotemberg, Kurtansky, Betz-Stablein, Caffery,
  Chousakos, Codella, Combalia, Dusza, Guitera, Gutman,
  et~al.]{rotemberg2021patient}
Rotemberg, V., Kurtansky, N., Betz-Stablein, B., Caffery, L., Chousakos, E.,
  Codella, N., Combalia, M., Dusza, S., Guitera, P., Gutman, D., et~al.
\newblock A patient-centric dataset of images and metadata for identifying
  melanomas using clinical context.
\newblock \emph{Scientific data}, 8\penalty0 (1):\penalty0 1--8, 2021.

\bibitem[Rudin(2019)]{rudin2019stop}
Rudin, C.
\newblock Stop explaining black box machine learning models for high stakes
  decisions and use interpretable models instead.
\newblock \emph{Nature Machine Intelligence}, 1\penalty0 (5):\penalty0
  206--215, 2019.

\bibitem[Sagawa et~al.(2019)Sagawa, Koh, Hashimoto, and Liang]{sagawa2019dro}
Sagawa, S., Koh, P.~W., Hashimoto, T.~B., and Liang, P.
\newblock Distributionally robust neural networks for group shifts: On the
  importance of regularization for worst-case generalization.
\newblock \emph{arXiv preprint arXiv:1911.08731}, 2019.

\bibitem[Samek et~al.(2016)Samek, Binder, Montavon, Lapuschkin, and
  M{\"u}ller]{samek2016evaluating}
Samek, W., Binder, A., Montavon, G., Lapuschkin, S., and M{\"u}ller, K.-R.
\newblock Evaluating the visualization of what a deep neural network has
  learned.
\newblock \emph{IEEE transactions on neural networks and learning systems},
  28\penalty0 (11):\penalty0 2660--2673, 2016.

\bibitem[Sarkar et~al.(2022)Sarkar, Vijaykeerthy, Sarkar, and
  Balasubramanian]{sarkar2021inducing}
Sarkar, A., Vijaykeerthy, D., Sarkar, A., and Balasubramanian, V.~N.
\newblock A framework for learning ante-hoc explainable models via concepts.
\newblock In \emph{Proceedings of the IEEE/CVF Conference on Computer Vision
  and Pattern Recognition}, pp.\  10286--10295, 2022.

\bibitem[Selvaraju et~al.(2017)Selvaraju, Cogswell, Das, Vedantam, Parikh, and
  Batra]{selvaraju2017grad}
Selvaraju, R.~R., Cogswell, M., Das, A., Vedantam, R., Parikh, D., and Batra,
  D.
\newblock Grad-cam: Visual explanations from deep networks via gradient-based
  localization.
\newblock In \emph{Proceedings of the IEEE international conference on computer
  vision}, pp.\  618--626, 2017.

\bibitem[Shrikumar et~al.(2016)Shrikumar, Greenside, Shcherbina, and
  Kundaje]{shrikumar2016not}
Shrikumar, A., Greenside, P., Shcherbina, A., and Kundaje, A.
\newblock Not just a black box: Learning important features through propagating
  activation differences.
\newblock \emph{arXiv preprint arXiv:1605.01713}, 2016.

\bibitem[Simonyan et~al.(2013)Simonyan, Vedaldi, and
  Zisserman]{simonyan2013deep}
Simonyan, K., Vedaldi, A., and Zisserman, A.
\newblock Deep inside convolutional networks: Visualising image classification
  models and saliency maps.
\newblock \emph{arXiv preprint arXiv:1312.6034}, 2013.

\bibitem[Singla et~al.(2019)Singla, Pollack, Chen, and
  Batmanghelich]{singla2019explanation}
Singla, S., Pollack, B., Chen, J., and Batmanghelich, K.
\newblock Explanation by progressive exaggeration.
\newblock \emph{arXiv preprint arXiv:1911.00483}, 2019.

\bibitem[Smilkov et~al.(2017)Smilkov, Thorat, Kim, Vi{\'e}gas, and
  Wattenberg]{smilkov2017smoothgrad}
Smilkov, D., Thorat, N., Kim, B., Vi{\'e}gas, F., and Wattenberg, M.
\newblock Smoothgrad: removing noise by adding noise.
\newblock \emph{arXiv preprint arXiv:1706.03825}, 2017.

\bibitem[Sundararajan et~al.(2017)Sundararajan, Taly, and
  Yan]{sundararajan2017axiomatic}
Sundararajan, M., Taly, A., and Yan, Q.
\newblock Axiomatic attribution for deep networks.
\newblock In \emph{International conference on machine learning}, pp.\
  3319--3328. PMLR, 2017.

\bibitem[Szegedy et~al.(2015)Szegedy, Liu, Jia, Sermanet, Reed, Anguelov,
  Erhan, Vanhoucke, and Rabinovich]{szegedy2015going}
Szegedy, C., Liu, W., Jia, Y., Sermanet, P., Reed, S., Anguelov, D., Erhan, D.,
  Vanhoucke, V., and Rabinovich, A.
\newblock Going deeper with convolutions.
\newblock In \emph{Proceedings of the IEEE conference on computer vision and
  pattern recognition}, pp.\  1--9, 2015.

\bibitem[Tschandl et~al.(2018)Tschandl, Rosendahl, and
  Kittler]{tschandl2018ham10000}
Tschandl, P., Rosendahl, C., and Kittler, H.
\newblock The ham10000 dataset, a large collection of multi-source
  dermatoscopic images of common pigmented skin lesions.
\newblock \emph{Scientific data}, 5\penalty0 (1):\penalty0 1--9, 2018.

\bibitem[Wadden et~al.(2019)Wadden, Wennberg, Luan, and
  Hajishirzi]{23_wadden-etal-2019-entity}
Wadden, D., Wennberg, U., Luan, Y., and Hajishirzi, H.
\newblock Entity, relation, and event extraction with contextualized span
  representations.
\newblock In \emph{Proceedings of the 2019 Conference on Empirical Methods in
  Natural Language Processing and the 9th International Joint Conference on
  Natural Language Processing (EMNLP-IJCNLP)}, pp.\  5784--5789, Hong Kong,
  China, November 2019. Association for Computational Linguistics.
\newblock \doi{10.18653/v1/D19-1585}.
\newblock URL \url{https://aclanthology.org/D19-1585}.

\bibitem[Wah et~al.(2011)Wah, Branson, Welinder, Perona, and
  Belongie]{wah2011caltech}
Wah, C., Branson, S., Welinder, P., Perona, P., and Belongie, S.
\newblock The caltech-ucsd birds-200-2011 dataset.
\newblock 2011.

\bibitem[Wan et~al.(2022)Wan, Belo, and Zejnilovic]{wan2022explainability}
Wan, C., Belo, R., and Zejnilovic, L.
\newblock Explainability's gain is optimality's loss? how explanations bias
  decision-making.
\newblock In \emph{Proceedings of the 2022 AAAI/ACM Conference on AI, Ethics,
  and Society}, pp.\  778--787, 2022.

\bibitem[Wang et~al.(2021)Wang, Yu, and Gao]{wang2021feature}
Wang, J., Yu, X., and Gao, Y.
\newblock Feature fusion vision transformer for fine-grained visual
  categorization.
\newblock \emph{arXiv preprint arXiv:2107.02341}, 2021.

\bibitem[Xian et~al.(2018)Xian, Lampert, Schiele, and Akata]{xian2018zero}
Xian, Y., Lampert, C.~H., Schiele, B., and Akata, Z.
\newblock Zero-shot learning—a comprehensive evaluation of the good, the bad
  and the ugly.
\newblock \emph{IEEE transactions on pattern analysis and machine
  intelligence}, 41\penalty0 (9):\penalty0 2251--2265, 2018.

\bibitem[Yeh et~al.(2019)Yeh, Kim, Arik, Li, Ravikumar, and
  Pfister]{yeh2019concept}
Yeh, C.-K., Kim, B., Arik, S., Li, C.-L., Ravikumar, P., and Pfister, T.
\newblock On concept-based explanations in deep neural networks.
\newblock 2019.

\bibitem[Yu et~al.(2022)Yu, Ghosh, Liu, Deible, and
  Batmanghelich]{yu2022anatomy}
Yu, K., Ghosh, S., Liu, Z., Deible, C., and Batmanghelich, K.
\newblock Anatomy-guided weakly-supervised abnormality localization in chest
  x-rays.
\newblock \emph{arXiv preprint arXiv:2206.12704}, 2022.

\bibitem[Yuksekgonul et~al.(2022)Yuksekgonul, Wang, and
  Zou]{yuksekgonul2022post}
Yuksekgonul, M., Wang, M., and Zou, J.
\newblock Post-hoc concept bottleneck models.
\newblock \emph{arXiv preprint arXiv:2205.15480}, 2022.

\bibitem[Zarlenga et~al.(2022)Zarlenga, Barbiero, Ciravegna, Marra, Giannini,
  Diligenti, Shams, Precioso, Melacci, Weller, et~al.]{zarlenga2022concept}
Zarlenga, M.~E., Barbiero, P., Ciravegna, G., Marra, G., Giannini, F.,
  Diligenti, M., Shams, Z., Precioso, F., Melacci, S., Weller, A., et~al.
\newblock Concept embedding models.
\newblock \emph{arXiv preprint arXiv:2209.09056}, 2022.

\end{thebibliography}
\bibliographystyle{icml2023}

\newpage
\appendix
\onecolumn
\section{Appendix}
\subsection{Project page }
Refer to the url \url{https://shantanu48114860.github.io/projects/ICML-2023-MoIE/} for the details of this project.
\label{app:code}

\subsection{Background of First-order logic (FOL) and Neuro-symbolic-AI}
\label{app:FOL}
FOL is a logical function
that accepts predicates (concept presence/absent) as input and returns a True/False output being a
logical expression of the predicates. The logical expression, which is a set of AND, OR, Negative,
and parenthesis, can be written in the so-called Disjunctive Normal Form (DNF)~\cite{mendelson2009introduction}. DNF is a FOL logical formula composed of a disjunction (OR) of conjunctions (AND), known as the ``sum of products''. 

Neuro-symbolic AI is an area of study that encompasses deep neural
networks with symbolic approaches to computing and AI to complement
the strengths and weaknesses of each, resulting in a robust AI capable
of reasoning and cognitive modeling~\cite{belle2020symbolic}.
Neuro-symbolic systems are hybrid models that leverage the robustness
of connectionist methods and the soundness of symbolic reasoning to
effectively integrate learning and reasoning
\cite{garcez2015neural,besold2017neural}.

\subsection{Learning the concepts}
\label{app:concept_learning}
As discussed in~\cref{sec:method}, $f^0: \mathcal{X} \rightarrow \mathcal{Y}$ is a pre-trained Blackbox. Also, $\displaystyle f^0(.) =  h^0 \circ \Phi(.)$. Here, $ \Phi: \mathcal{X} \rightarrow R^l $ is the image embeddings, transforming the input images to an intermediate representation and $ h^0: R^l \rightarrow \mathcal{Y}$ is the classifier, classifying the output $\mathcal{Y}$ using the embeddings, $\Phi$. Our approach is applicable for both datasets with and without human-interpretable concept annotations. For datasets with the concept annotation $\mathcal{C} \in \mathbb{R}^{N_c}$ ($N_c$ being the number of concepts per image $\mathcal{X}$), we learn $t: R^l \rightarrow\mathcal{C}$ to classify the concepts using the embeddings. Per this definition, $t$ outputs a scalar value $c$ representing a single concept for each input image. 
We adopt the concept learning strategy in PosthocCBM (PCBM)~\cite{yuksekgonul2022post} for datasets without concept annotation. 
Specifically, we leverage a set of image embeddings with the concept being present and absent. Next, we learn a linear SVM ($t$) to construct the concept activation matrix~\cite{kim2017interpretability} as $\boldsymbol{Q} \in\mathbb{R}^{N_c \times l}$. 
Finally we estimate the concept value as $c = \frac{<\Phi(x), q^i>}{||q_i||_2^2}$ $ \in \mathbb{R}$ utilizing each row $\boldsymbol{q^i}$ of $\boldsymbol{Q}$. Thus, the complete tuple of $j^{th}$ sample is $\{x_j, y_j, c_j\}$, denoting the image, label, and learned concept vector, respectively.

\subsection{Optimization}
\label{app:loss}
In this section, we will discuss the loss function used in distilling the knowledge from the blackbox to the symbolic model. We remove the superscript $k$ for brevity. We adopted the optimization proposed in \cite{geifman2019selectivenet}.Specifically, we convert the constrained optimization problem in~\cref{equ: optimization_g} as 

\begin{align}
\label{equ:unconstrained_risk}
&\mathcal{L}_s = \mathcal{R}(\pi, g) + \lambda_s \Psi(\tau - \zeta(\pi))\\ \nonumber
&\Psi(a) = \text{max}(0, a)^2 ,
\end{align}

where $\tau$ is the target coverage and $\lambda_s$ is a hyperparameter (Lagrange multiplier). We define $\mathcal{R}(.)$ and $\mathcal{L}_{g, \pi}(.)$ in~\cref{equ: emp_risk} and~\cref{equ: g_k} respectively. $\ell$ in~\cref{equ: g_k} is defined as follows:

\begin{align}
\label{equ:ell}
\ell\big(f, g \big) &= \ell_{distill}(f, g) + \lambda_{lens}\sum_{i=1}^r\mathcal{H}(\beta^i) ,
\end{align}

where $\lambda_{lens}$ and $\mathcal{H}(\beta^i)$ are the hyperparameters and entropy regularize, introduced in \cite{barbiero2022entropy} with $r$ being the total number of class labels. Specifically, $\beta^i$ is the categorical distribution of the weights corresponding to each concept.  To select only a few relevant concepts for each target class, higher values of $\lambda_{lens}$ will lead to a sparser configuration of $\beta$. $\ell$ is the knowledge distillation loss \cite{hinton2015distilling}, defined as 

\begin{align}
\label{equ:distill}
\ell(f, g) = & (\alpha_{KD}* T_{KD}*T_{KD}) KL\big(\text{LogSoftmax}(g(.)/T_{KD}) , \text{Softmax}(f(.)/T_{KD})\big) + \\ \nonumber
& (1 - \alpha_{KD}) CE\big(g(.), y\big),
\end{align}

where $T_{KD}$ is the temperature, CE is the Cross-Entropy loss, and $\alpha_{KD}$ is relative weighting controlling the supervision from the blackbox $f$ and the class label $y$.

As discussed in \cite{geifman2019selectivenet}, we also define an auxiliary interpretable model using the same prediction task assigned to $g$ using the following loss function

\begin{align}
\label{equ:aux}
\mathcal{L}_{aux} = \frac{1}{m}\sum_{j=1}^m\ell_{distill}(f(\boldsymbol{x_j}), g(\boldsymbol{c_j})) + \lambda_{lens}\sum_{i=1}^r\mathcal{H}(\beta^i),
\end{align}
which is agnostic of any coverage. $\mathcal{L}_{aux}$ is necessary for optimization as the symbolic model will focus on the target coverage $\tau$ before learning any relevant features, overfitting to the wrong subset of the training set. The final loss function to optimize by g in each iteration is as follows:

\begin{align}
\label{equ:final_loss_g}
\mathcal{L} = \alpha \mathcal{L_s} + (1 - \alpha)\mathcal{L}_{aux},
\end{align}

where $\alpha$ is the can be tuned as a hyperparameter. Following \cite{geifman2019selectivenet}, we also use $\alpha=0.5$ in all of our experiments.

\subsection{Algorithm}
\label{app:algo}

\begin{figure}[h]
\centering
\includegraphics[width=1\textwidth]
{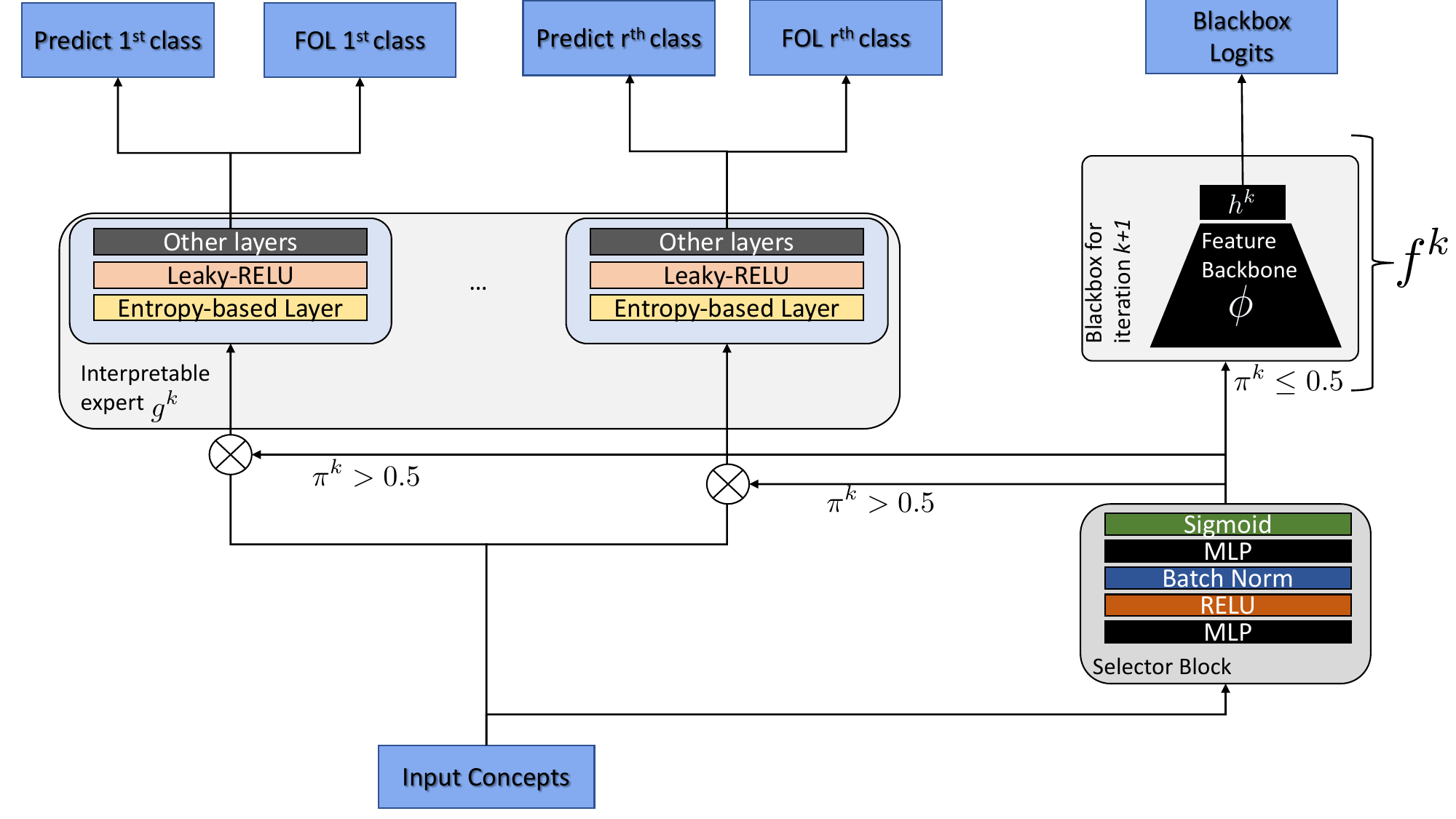}
\caption{Architecture of MoIE. In an iteration $k$ during inference, the selector routes the samples to go through the interpretable expert $g^k$ if the probability $\pi^k \ge 0.5$. If $\pi^k < 0.5$, the selector routes the samples, through $f^k$, the Blackbox for iteration $k+1$. Note $f^k = h^k(\Phi(.)$ is an approximation of the residual $r^k = f^{k-1} - g^k$.  }
\label{fig:architecture}
\end{figure}

\begin{algorithm}[h]
   \caption{\emph{Route, interpret} and \emph{repeat} algorithm to generate FOL explanations locally.}
   \label{algo: train}
\begin{algorithmic}[1]
   \STATE {\bfseries Input:} Complete tuple: \{$x_j$, $y_j$, $c_j$\}$_{j=1}^n$; initial blackbox $f^0 = h^0(\Phi(.))$; K as the total iterations; Coverages $\tau_1, \dots ,\tau_K$.
   \STATE {\bfseries Output:} Sparse mixture of experts and their selectors $\{g^k, \pi^k\}_{k=1}^K$ and the final residual $f^K = h^K(\Phi(.))$
   \STATE Fix $\Phi$.
   \FOR{\texttt{$k=1 \dots K $}}
       \STATE  Fix $\pi^1 \dots \pi^{k-1}$.
       \STATE Minimize $\mathcal{L}^k$ using equation \ref{equ:final_loss_g} to learn $\pi^k$ and $g^k$.
       \STATE Calculate $r^k = f^{k-1}(.) - g^k(.)$
       \STATE Minimize equation \ref{equ: residual} to learn $f^k(.)$, the new blackbox for the next iteration $k+1$.
       \ENDFOR
        \FOR{\texttt{$k=1 \dots K $}}
            \FOR{sample $j$ in \texttt{test-set}}
                \REPEAT
                    \STATE Initialize \texttt{sub\_select\_concept} = $True$
                    \STATE Initialize the \texttt{percentile\_threshold} = $99$.
                    \STATE Retrieve the predicted class label of sample $j$ from the expert $k$ as: $\hat{y}_j = g^k(c_j)$
                    \STATE Create a mask vector $m_j$. $m_j[i] = 1$ if 
                    $\Tilde{\alpha}[\hat{y}_j][i] \geq$ percentile$(\Tilde{\alpha}[\hat{y}_j]$, \texttt{percentile\_threshold}$)$ and $0$ otherwise. Specifically, the $i^{th}$ entry in $m_j$ is one if the $i^{th}$ value of the attention score $\Tilde{\alpha}[\hat{y}_j]$ is greater than (\texttt{percentile\_attention})$^{th}$ percentile. 
                    \STATE Subselect the concept vector as $\Tilde{c}_j$ as: $\Tilde{c}_j = c_j\odot m_j$
                    \IF{$g^k(\Tilde{c}_j) \neq \hat{y}_j$}
                        \STATE \texttt{percentile\_threshold} = \texttt{percentile\_threshold} - 1
                        \STATE \texttt{sub\_select\_concept} = $false$
                    \ENDIF
                \UNTIL \texttt{sub\_select\_concept} is $True$
                \STATE Using the subselected concept vector $\Tilde{c_j}$, construct the FOL expression of the $j^{th}$ sample as suggested by~\cite{barbiero2022entropy}.
            \ENDFOR
    \ENDFOR
\end{algorithmic}
\end{algorithm}

\cref{algo: train} explains the overall training procedure of our method.~\cref{fig:architecture} displays the architecture of our model in iteration $k$.

\subsection{Dataset}
\label{app:dataset}
\paragraph{CUB-200}
The Caltech-UCSD Birds-200-2011 (\cite{wah2011caltech}) is a fine-grained classification dataset comprising 11788 images and 312 noisy visual concepts. The aim is to classify the correct bird species from 200 possible classes. We adopted the strategy discussed in~\cite{barbiero2022entropy} to extract 108 denoised visual concepts. Also, we utilize training/validation splits shared in \cite{barbiero2022entropy}. Finally, we use the state-of-the-art classification models Resnet-101 (\cite{he2016deep}) and Vision-Transformer (VIT) (\cite{wang2021feature}) as the blackboxes $f^0$. 


\paragraph{Animals with attributes2 (Awa2)}
 AwA2 dataset \cite{xian2018zero} consists of 37322 images of total 50 animals classes with 85 numeric attribute. We use the state-of-the-art classification models Resnet-101 (\cite{he2016deep}) and Vision-Transformer (VIT) (\cite{wang2021feature}) as the blackboxes $f^0$.

\paragraph{HAM10000}
HAM10000 (\cite{tschandl2018ham10000}) is a classification dataset aiming to classify a skin lesion benign or malignant. Following \cite{daneshjou2021disparities}, we use Inception \cite{szegedy2015going} model, trained on this dataset as the blackbox $f^0$. We follow the strategy in \cite{lucieri2020interpretability} to extract the 8 concepts from the Derm7pt (\cite{kawahara2018seven}) dataset.

\paragraph{SIIM-ISIC}
To test a real-world transfer learning use case, we evaluate the
model trained on HAM10000 on a subset of the SIIM-ISIC\cite{rotemberg2021patient}) Melanoma Classification dataset. We use
the same concepts described in the HAM10000 dataset.

\paragraph{MIMIC-CXR} We use  220,763 frontal images from the MIMIC-CXR dataset \cite{12_johnsonmimic} aiming to classify effusion. We obtain the anatomical and observation concepts from the RadGraph annotations in RadGraph’s inference dataset (\cite{10_jain2021radgraph}), automatically generated by DYGIE++ (\cite{23_wadden-etal-2019-entity}). We use the test-train-validation splits from \cite{yu2022anatomy} and Densenet121 \cite{huang2017densely} as the blackbox $f^0$.

\subsection{Estimation of completeness score}
\label{app:completeness}

Let $f^0(x)=h^0(\Phi(\boldsymbol{x})$ is the initial Blackbox as per~\cref{sec:method}. The Concept completeness paper~\cite{yeh2019concept} assumes $\Phi(\boldsymbol{x}) \in \mathbb{R}^l$ (\st $l=T.d$) to be a concatanation of $[\phi(\boldsymbol{x}_1), \phi(\boldsymbol{x}_2), \dots, \phi(\boldsymbol{x}_T)]$ \st $\phi(\boldsymbol{x}) \in \mathbb{R}^d$. Recall we utilize $t$ to learn the concepts $\mathcal{C}$ with $N_c$ being the total number of concepts per image. So the parameters of $t$, represented by $\omega_1, \omega_2, \dots \omega_{N_c}$ \st $\omega_i \in \mathbb{R}^d$ represent linear direction in the embedding space $\phi(.) \in \mathbb{R}^d$. Next, we compute the concept product $v_c(\boldsymbol{x}_t)(<\phi(\boldsymbol{x}_t), \omega_j>)_{j=1}^{N_c}$, denoting the similarity between the image embedding and linear direction of $j^{th}$ concept. Finally, we normalize $v_c(.)$ to obtain the concept score as 
$v_v(\boldsymbol{x}) = \big(\frac{v_c(\boldsymbol{x_t})}{||v_c(\boldsymbol{x}_t)||_2}\big)_{t=1}^T \in \mathbb{R}^{T.{N_c}}$.

Next for a Blackbox $f^0(x)=h^0(\Phi(\boldsymbol{x})$, set of concepts $c_1, c_2, \dots c_{N_c}$ and their linear direction  $\omega_1, \omega_2, \dots \omega_{N_c}$ in the embedding space and, we compute the completeness score as:

\begin{align}
\eta_{f^0} = \frac{\text{sup}_\Gamma \mathbb{P}_{\boldsymbol{x}, y \sim V}
[y = \operatorname*{arg\,max}_{y'}h^0_{y'}(\Gamma(v_c(\boldsymbol{x})))] - a_r}{
\mathbb{P}_{\boldsymbol{x}, y \sim V}
[y = \operatorname*{arg\,max}_{y'}f^0_{y'}(\boldsymbol{x})] - a_r
},
\end{align}

where $V$ is the validation set and $\Gamma : \mathbb{R}^{T.m} \rightarrow \mathbb{R}^l$, projection from the concept score to the embedding space$\Phi$. For CUB-200 and Awa2 we estimate $\mathbb{P}_{\boldsymbol{x}, y \sim V}
[y = \operatorname*{arg\,max}_{y'}h^0_{y'}(\Gamma(v_c(\boldsymbol{x})))]$ as the best accuracy using the given concepts and $a_r$ is the random accuracy. For HAM10000, we estimate the same as the best AUROC. Completeness score indicates the consistency between the prediction based just on concepts and the given Blackbox$f^0$. If the identified concepts are sufficiently rich, label prediction will be similar to the Blackbox, resulting in higher completeness scores for the concept set. In all our experiments, $\Gamma$ is a two-layer feedforward neural network with 1000 neurons.

To plot the completeness score in~\cref{fig:valid_concepts}a-c, we select the topN concepts iteratively representing the $N < N_c$ concepts most significant to the prediction of the interpretable model $g$. Recall we follow Entropy based linear neural network~\cite{barbiero2022entropy} as $g$. So each concept has an associated attention score, $\alpha$ in $g$~\cite{barbiero2022entropy}, denoting the importance of the concept for the prediction. We select the topN concepts based on the $N$ concepts with highest attention weights. We get the linear direction of these topN concepts from the parameters of the learned $t$ and project it to the embedding space $\Phi$ using $\Gamma$. If $\Gamma$ reconstructs the discriminative features from the concepts successfully, the concepts achieves high completeness scores, showing faithfulness with the Blackbox. Recall~\cref{fig:valid_concepts}a-c demonstrate that MoIE outperforms the baselines in terms of the completeness scores. This suggests that MoIE identifies rich instance-specific concepts than the baselines, being consistent with the Blackbox.

\subsection{Architectural details of symbolic experts and hyperparameters}
\label{app:g}

\cref{tab:bb_config} demonstrates different settings to train the Blackbox of CUB-200, Awa2 and MIMIC-CXR respectively. For the VIT-based backbone, we used the same hyperparameter setting used in the state-of-the-art Vit-B\_16 variant in \cite{wang2021feature}. To train $t$, we flatten the feature maps from the last convolutional block of $\Phi$ using ``Adaptive average pooling'' for CUB-200 and Awa2 datasets.For MIMIC-CXR and HAM10000, we flatten out the feature maps from the last convolutional block. For VIT-based backbones, we take the first block of representation from the encoder of VIT.  For HAM10000, we use the same Blackbox in \cite{yuksekgonul2022post}.~\cref{tab:g_config_cub_200},~\cref{tab:g_config_awa2},~\cref{tab:g_config_ham10k},~\cref{tab:g_config_mimic_cxr} enumerate all the different settings to train the interpretable experts for CUB-200, Awa2, HAM, and MIMIC-CXR respectively. All the residuals in different iterations follow the same settings as their blackbox counterparts.

\begin{table}[h]
\caption{Hyperparameter setting of different convolution-based Blackboxes used by CUB-200, Awa2 and MIMIC-CXR}
\label{tab:bb_config}
\begin{center}
\begin{tabular}{l c c c}
\toprule 
     {\textbf{Setting}} & {\textbf{CUB-200}} & {\textbf{Awa2}} & 
    {\textbf{MIMIC-CXR}}\\
\midrule 
       Backbone              & ResNet-101 & ResNet-101 & DenseNet-121  \\
       Pretrained on ImageNet      & True &True & True \\
       Image size            & 448 & 224 & 512 \\
       Learning rate         & 0.001 & 0.001 & 0.01 \\
       Optimization          & SGD & Adam & SGD \\
       Weight-decay      & 0.00001 & 0 & 0.0001 \\
       Epcohs             & 95 & 90 & 50 \\
       Layers used as $\Phi$ &  \makecell{till 4$^{th}$ ResNet \\Block} &  \makecell{till 4$^{th}$ ResNet \\Block} &  \makecell{till 4$^{th}$ DenseNet \\Block} \\
       Flattening type for the input to $t$    &  \makecell{Adaptive average \\pooling} &  \makecell{Adaptive average \\pooling} & Flatten \\
\bottomrule
\end{tabular}
\end{center}
\end{table}

\begin{table}[h]
\caption{Hyperparameter setting of interpretable experts ($g$) trained on ResNet-101 (top) and VIT (bottom) blackboxes for the CUB-200 dataset}
\label{tab:g_config_cub_200}
\begin{center}
\begin{tabular}{l|c|c|c|c|c|c}
\toprule 
    \thead{\textbf{Settings based on dataset}} & \thead{\textbf{Expert1}} & \thead{\textbf{Expert2}} 
    & \thead{\textbf{Expert3}} & \thead{\textbf{Expert4}} & \thead{\textbf{Expert5}} & \thead{\textbf{Expert6}}\\
\midrule 
        CUB-200 (ResNet-101)              &    &   &  & &  & \\
       \quad + Batch size              & 16 & 16 & 16 & 16 & 16 & 16   \\
        
       \quad + Coverage ($\tau$)  & 0.2 & 0.2 & 0.2 & 0.2 & 0.2 & 0.2 \\
       
       \quad + Learning rate & 0.01 & 0.01 & 0.01 & 0.01 & 0.01 & 0.01 \\
       
       \quad + $\lambda_{lens}$ & 0.0001 & 0.0001 & 0.0001 & 0.0001 & 0.0001 & 0.0001 \\
    
       \quad +$\alpha_{KD}$ & 0.9 & 0.9 & 0.9 & 0.9 & 0.9 & 0.9 \\
       \quad + $T_{KD}$ & 10 & 10 & 10 & 10 &10 & 10 \\
       \quad +hidden neurons & 10 & 10 & 10 & 10 &10 & 10 \\
       \quad +$\lambda_s$ & 32 & 32 & 32 & 32 & 32 & 32 \\
       \quad + $T_{lens}$ & 0.7 & 0.7 & 0.7 & 0.7 & 0.7 & 0.7 \\
\midrule 
        CUB-200 (VIT)            &    &   &  & &  & \\
       \quad + Batch size              & 16 & 16 & 16 & 16 & 16 & 16   \\
        
       \quad + Coverage ($\tau$)  & 0.2 & 0.2 & 0.2 & 0.2 & 0.2 & 0.2 \\
       
       \quad + Learning rate & 0.01 & 0.01 & 0.01 & 0.01 & 0.01 & 0.01 \\

       \quad + $\lambda_{lens}$ & 0.0001 & 0.0001 & 0.0001 & 0.0001 & 0.0001 & 0.0001 \\
    
       \quad +$\alpha_{KD}$ & 0.99 & 0.99 & 0.99 & 0.99 & 0.99 & 0.99 \\
       \quad + $T_{KD}$ & 10 & 10 & 10 & 10 &10 & 10 \\
       \quad +hidden neurons & 10 & 10 & 10 & 10 &10 & 10 \\
       \quad +$\lambda_s$ & 32 & 32 & 32 & 32 & 32 & 32 \\
       \quad +$T_{lens}$ & 6.0 & 6.0 & 6.0 & 6.0 & 6.0 & 6.0 \\
\bottomrule
\end{tabular}
\end{center}
\end{table}

\begin{table}[h]
\caption{Hyperparameter setting of interpretable experts ($g$) trained on ResNet-101 (top) and VIT (bottom) blackboxes for the Awa2 dataset}
\label{tab:g_config_awa2}
\begin{center}
\begin{tabular}{l|c|c|c|c|c|c}
\toprule 
    \thead{\textbf{Settings based on dataset}} & \thead{\textbf{Expert1}} & \thead{\textbf{Expert2}} 
    & \thead{\textbf{Expert3}} & \thead{\textbf{Expert4}} & \thead{\textbf{Expert5}} & \thead{\textbf{Expert6}}\\
\midrule 
        Awa2 (ResNet-101)              &    &   &  & &  & \\
       \quad + Batch size              & 30 & 30 & 30 & 30 & - & -   \\
        
       \quad + Coverage ($\tau$)  & 0.4 & 0.35 & 0.35 & 0.25 & - & - \\
       
       \quad + Learning rate & 0.001 & 0.001 & 0.001 & 0.001 & - & - \\
       
       \quad + $\lambda_{lens}$ & 0.0001 & 0.0001 & 0.0001 & 0.0001 & - & - \\
    
       \quad +$\alpha_{KD}$ & 0.9 & 0.9 & 0.9 & 0.9 & - & - \\
       \quad + $T_{KD}$ & 10 & 10 & 10 & 10 & - & - \\
       \quad +hidden neurons & 10 & 10 & 10 & 10 & - & - \\
       \quad +$\lambda_s$ & 32 & 32 & 32 & 32 & - & - \\
       \quad + $T_{lens}$ & 0.7 & 0.7 & 0.7 & 0.7 & - & - \\
\midrule 
        Awa2 (VIT)            &    &   &  & &  & \\
       \quad + Batch size              & 30 & 30 & 30 & 30 & 30 & 30   \\
        
       \quad + Coverage ($\tau$)  & 0.2 & 0.2 & 0.2 & 0.2 & 0.2 & 0.2 \\
       
       \quad + Learning rate & 0.01 & 0.01 & 0.01 & 0.01 & 0.01 & 0.01 \\

       \quad + $\lambda_{lens}$ & 0.0001 & 0.0001 & 0.0001 & 0.0001 & 0.0001 & 0.0001 \\
    
       \quad +$\alpha_{KD}$ & 0.99 & 0.99 & 0.99 & 0.99 & 0.99 & 0.99 \\
       \quad + $T_{KD}$ & 10 & 10 & 10 & 10 &10 & 10 \\
       \quad +hidden neurons & 10 & 10 & 10 & 10 &10 & 10 \\
       \quad +$\lambda_s$ & 32 & 32 & 32 & 32 & 32 & 32 \\
       \quad + $T_{lens}$ & 6.0 & 6.0 & 6.0 & 6.0 & 6.0 & 6.0 \\
\bottomrule
\end{tabular}
\end{center}
\end{table}

\begin{table}[h]
\caption{Hyperparameter setting of interpretable experts ($g$) for the dataset HAM10000}
\label{tab:g_config_ham10k}
\begin{center}
\begin{tabular}{l|c|c|c|c|c|c}
\toprule 
    \thead{\textbf{Settings based on dataset}} & \thead{\textbf{Expert1}} & \thead{\textbf{Expert2}} 
    & \thead{\textbf{Expert3}} & \thead{\textbf{Expert4}} & \thead{\textbf{Expert5}}
    & {\textbf{Expert6}}\\
\midrule 
        HAM10000 (Inception-V3)              &    &   &  & & &  \\
       \quad + Batch size              & 32 & 32 & 32 & 32 & 32&  32   \\
        
       \quad + Coverage ($\tau$)  & 0.4 & 0.2 & 0.2 & 0.2 & 0.1&  0.1\\
       
       \quad + Learning rate & 0.01 & 0.01 & 0.01 & 0.01 & 0.01& 0.01 \\
       
       \quad + $\lambda_{lens}$ & 0.0001 & 0.0001 & 0.0001 & 0.0001 & 0.0001 &  0.0001\\
    
       \quad +$\alpha_{KD}$ & 0.9 & 0.9 & 0.9 & 0.9 & 0.9& 0.9\\
       \quad + $T_{KD}$ & 10 & 10 & 10 & 10 & 10& 10 \\
       \quad +hidden neurons & 10 & 10 & 10 & 10 & 10& 10\\
       \quad +$\lambda_s$ & 64 & 64 & 64 & 64 & 64& 64  \\
       \quad + $T_{lens}$ & 0.7 & 0.7 & 0.7 & 0.7 & 0.7& 0.7 \\
\bottomrule
\end{tabular}
\end{center}
\end{table}

\begin{table}[H]
\caption{Hyperparameter setting of interpretable experts ($g$) for the dataset MIMIC-CXR}
\label{tab:g_config_mimic_cxr}
\begin{center}
\begin{tabular}{l|c|c|c}
\toprule 
    \thead{\textbf{Settings based on dataset}} & \thead{\textbf{Expert1}} & \thead{\textbf{Expert2}} 
    & \thead{\textbf{Expert3}} \\
\midrule 
        Effusion-MIMIC-CXR (DenseNet-121)              &    &   &     \\
       \quad + Batch size              & 1028 & 1028 & 1028     \\
        
       \quad + Coverage ($\tau$)  & 0.6 & 0.2 & 0.15   \\
       
       \quad + Learning rate & 0.01 & 0.01 & 0.01 \\
       
       \quad + $\lambda_{lens}$ & 0.0001 & 0.0001 & 0.0001  \\
    
       \quad +$\alpha_{KD}$ & 0.99 & 0.99 & 0.99  \\
       \quad + $T_{KD}$ & 20 & 20 & 20   \\
       \quad +hidden neurons & 20, 20 & 20, 20 & 20, 20  \\
       \quad +$\lambda_s$ & 96 & 128 & 256   \\
       \quad +$T_{lens}$ & 7.6 & 7.6 & 7.6 \\
\bottomrule
\end{tabular}
\end{center}
\end{table}

\subsection{Flow diagram to eliminate shotcut}
\label{app:shortcut}
\begin{figure*}[h]
\centering
\includegraphics[width=1.0\textwidth]
{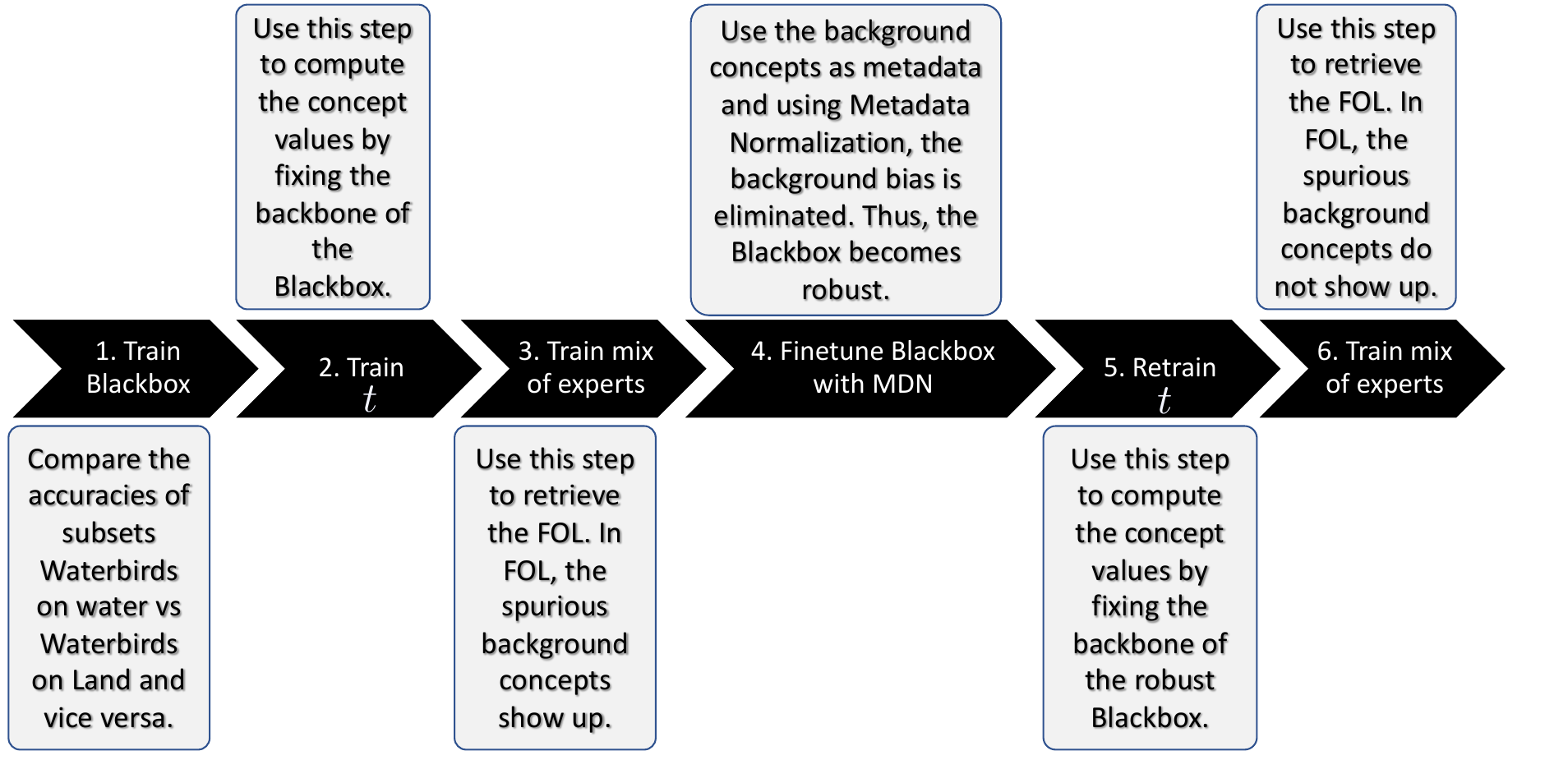}
\caption{The flow diagram to eliminate the shortcut from vision datasets using FOL by MoIE.}
\label{fig:spurious_flow}
\end{figure*}

\cref{fig:spurious_flow} shows the flow digram to eliminate shortcut.

\subsection{More Results}
\subsubsection{Comparison with other interpretable by design baselines}
\label{app:more_baselines}
\begin{table}[H]
\caption{Comparing the performance of MoIE with additional interpretable by design baselines.
}
\fontsize{7.5pt}{0.20cm}\selectfont
\label{tab:performance_app}
\begin{center}
\begin{tabular}{p{23em} c c c c}
\toprule 
        \textbf{MODEL} & \multicolumn{3}{c}{\textbf{DATASET}} \\
       & CUB-200 (RESNET101) & AWA2 (RESNET101) & EFFUSION  \\
\midrule 
    BLACKBOX & 0.88 & 0.89 & 0.91 \\
\midrule 
ANTEHOC W SUP~\cite{sarkar2021inducing} & 0.71 & 0.85 & 0.75\\
ANTEHOC W/O SUP~\cite{sarkar2021inducing} & 0.64 & 0.81 & 0.70\\
HARD W AR~\cite{havasi2022addressing} & 0.81 & 0.84 & 0.73\\
HARD W/O AR~\cite{havasi2022addressing} & 0.78 & 0.81 & 0.71\\
\midrule
\textbf{OURS} \\
MoIE (COVERAGE) &\textbf{0.86} &
     \textbf{0.87}
     & \textbf{0.87} \\
MoIE + RESIDUAL &\textbf{0.84} &
     \textbf{0.86}
     & \textbf{0.86} \\
\bottomrule
\end{tabular}
\end{center}
\end{table}
~\cref{tab:performance_app} compares our method with several other interpretable by design baselines.

\subsubsection{Results of Effusion of MIMIC-CXR}
\label{app:mimic_cxr}
\begin{figure*}[h]
\centering
\includegraphics[width=1.0\textwidth]
{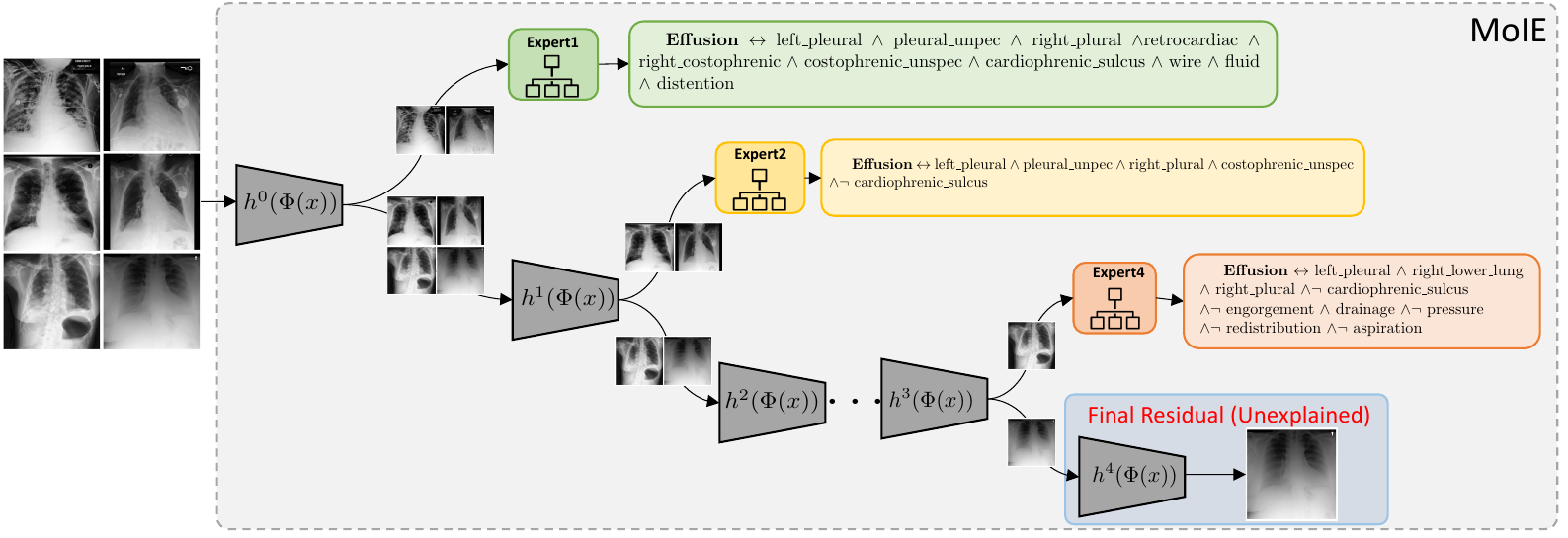}
\caption{Construction logical explanations of the samples of ``Effusion'' in the MIMIC-CXR dataset for various experts in MoIE at inference. The final residual covers the unexplained sample, which is ``harder'' to explain (indicated in \emph{red}).}
\label{fig:mimic_concept_explanation}
\end{figure*}

\begin{figure*}[h]
\centering
\includegraphics[width=1.0\textwidth]
{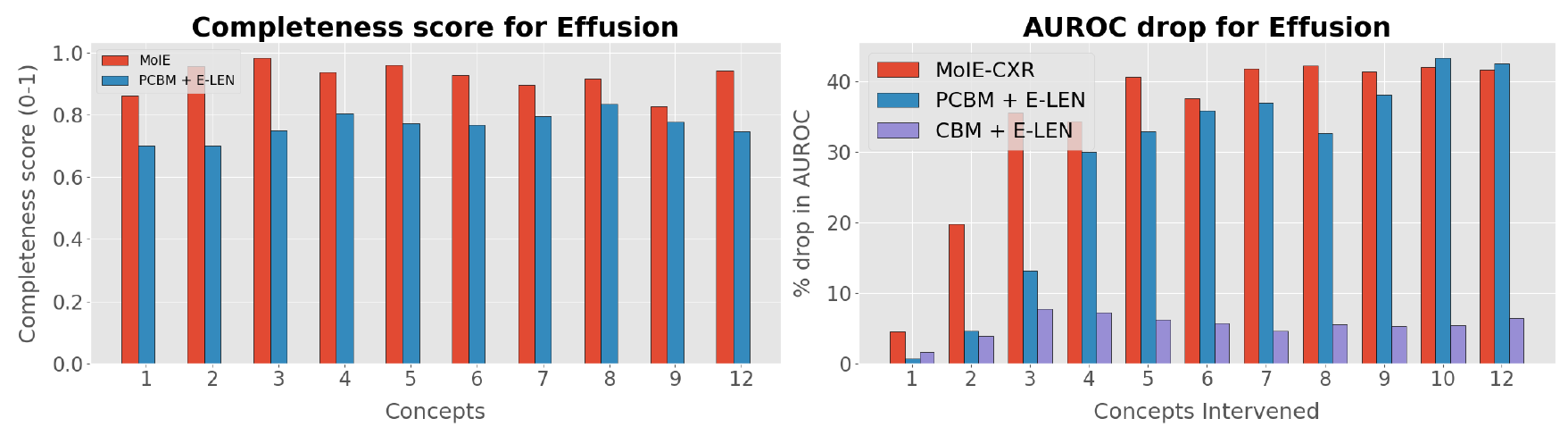}
\caption{\textbf{(a)} Completeness scores for different significant concepts of Effusion.  \textbf{(b)} Drop in AUROC by zeroing out the concepts for Effusion.}
\label{fig:mimic_concept_quant}
\end{figure*}

\begin{figure*}[h]
\centering
\includegraphics[width=1.0\textwidth]
{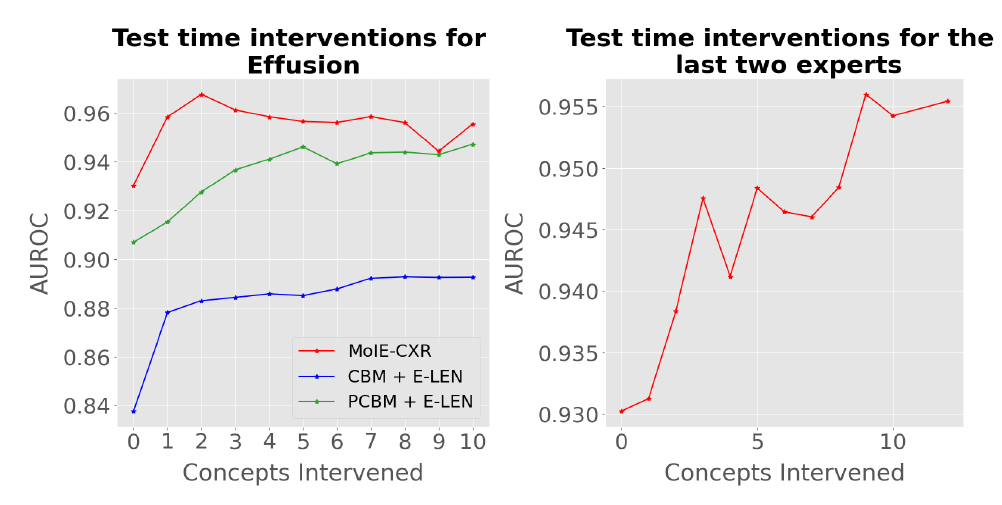}
\caption{\textbf{(a)} Test time interventions of different concepts on all samples for Effusion.
\textbf{(b)} Test time interventions of different concepts on only the ``hard'' samples covered by the last two experts for Effusion}
\label{fig:mimic_concept_tti}
\end{figure*}

~\cref{fig:mimic_concept_explanation} demonstrates the diversity of instance-specific local FOL explanations of different concepts of MoIE and the final residual.~\cref{fig:mimic_concept_quant}(a) shows the completeness scores for different concepts.~\cref{fig:mimic_concept_quant}(b) shows the drop in AUROC while zeroing out different concepts. ~\cref{fig:mimic_concept_tti}(a) shows test time interventions of different concepts on all samples. ~\cref{fig:mimic_concept_tti}(b) shows test time interventions of different concepts on only the ``hard'' samples covered by the last two experts.

\subsubsection{Performance of experts and residual for ResNet-derived experts of Awa2 and CUB-200 datasets}
\label{app:resnet_cv}
\begin{figure}[h]
\centering
\includegraphics[width=1\linewidth]{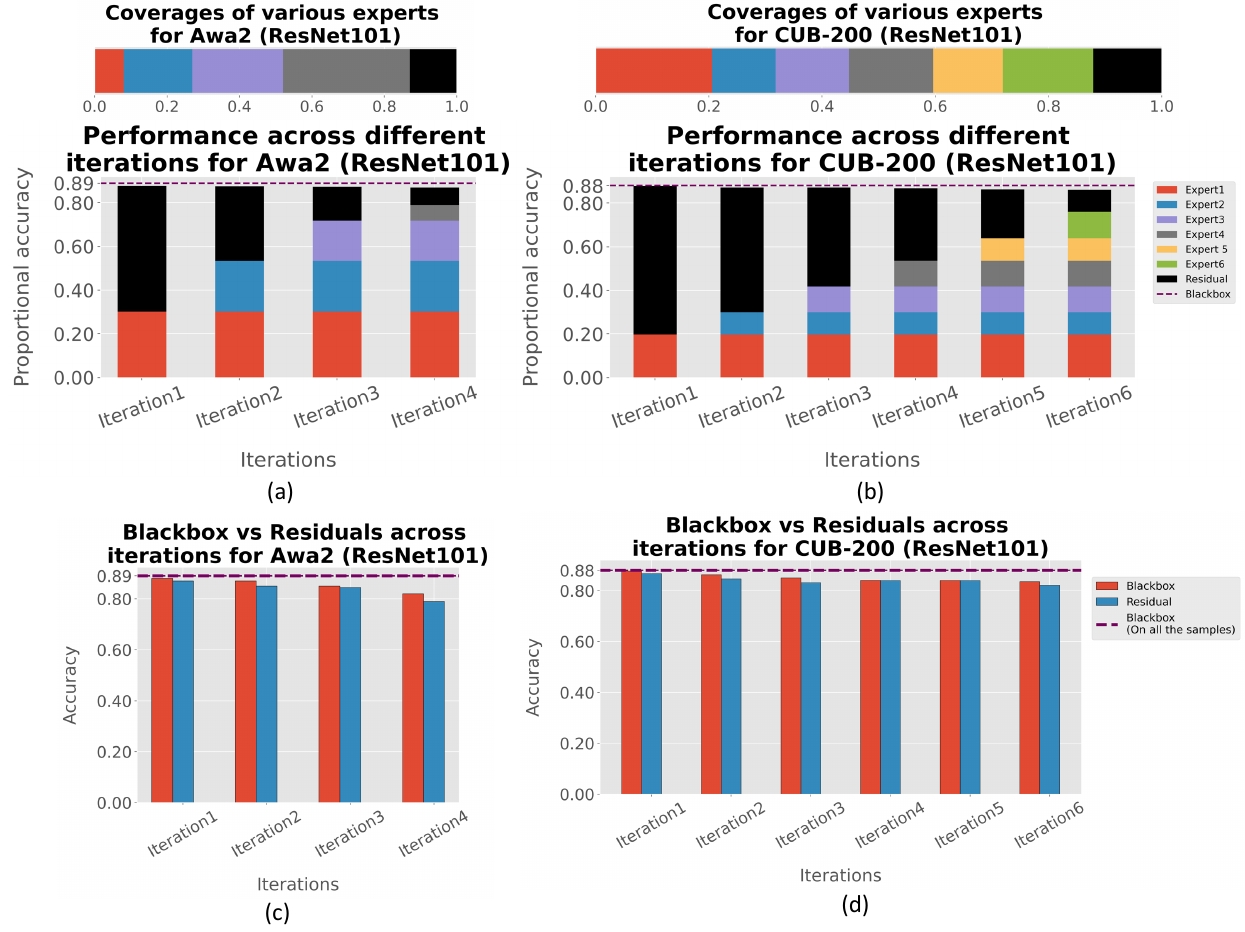}
\caption{The performances of experts and residuals across iterations for ResNet derived MoIE for CUB-200 and Awa2. 
\textbf{(a-b)} Coverage and proportional accuracy of the experts and residuals. 
\textbf{(c-d)} We route the samples covered by the residuals across iterations to the initial Blackbox $f^0$ and compare the accuracy of $f^0$ (red bar) with the residual (blue bar).}
\label{fig:expert_performance_cv_resnet}
\end{figure}

\cref{fig:expert_performance_cv_resnet} shows the coverage (top row), performances (bottom row) of each expert and residual across iterations of - (a) ResNet101-derived Awa2 and (b) ResNet101-derived CUB-200 respectively.

\subsubsection{Concept validation of Awa2}
\label{app:awa2}
\begin{figure}
\centering
\includegraphics[width=\columnwidth]
{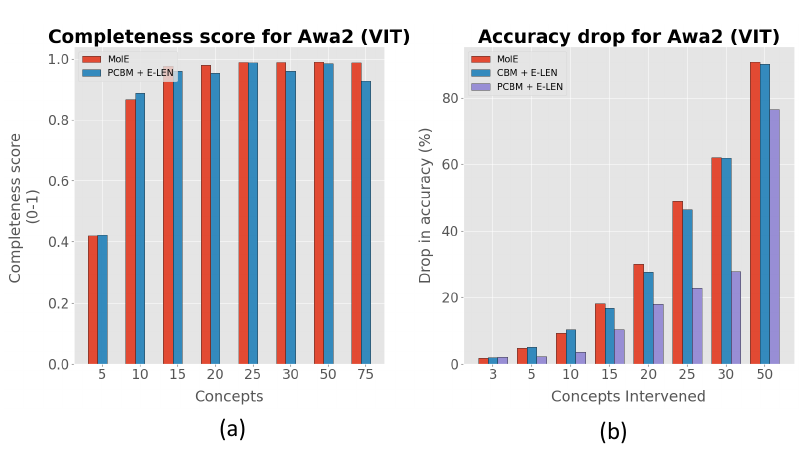}
\vskip -7pt
\caption{\textbf{(a)} Completeness scores for different significant concepts of Awa2.  \textbf{(b)} Drop in accuracy by zeroing out the concepts for Awa2.}
\vskip -10pt
\label{fig:completeness_acc_awa2}
\end{figure}

\cref{fig:completeness_acc_awa2} shows the completeness scores and the drop in accuracy by zeroing out the concepts for Awa2.




\subsubsection{Example of expert-specific test time intervention}
\label{app:tti_qual}
\begin{figure}[h]
\centering
\includegraphics[width=1 \linewidth]{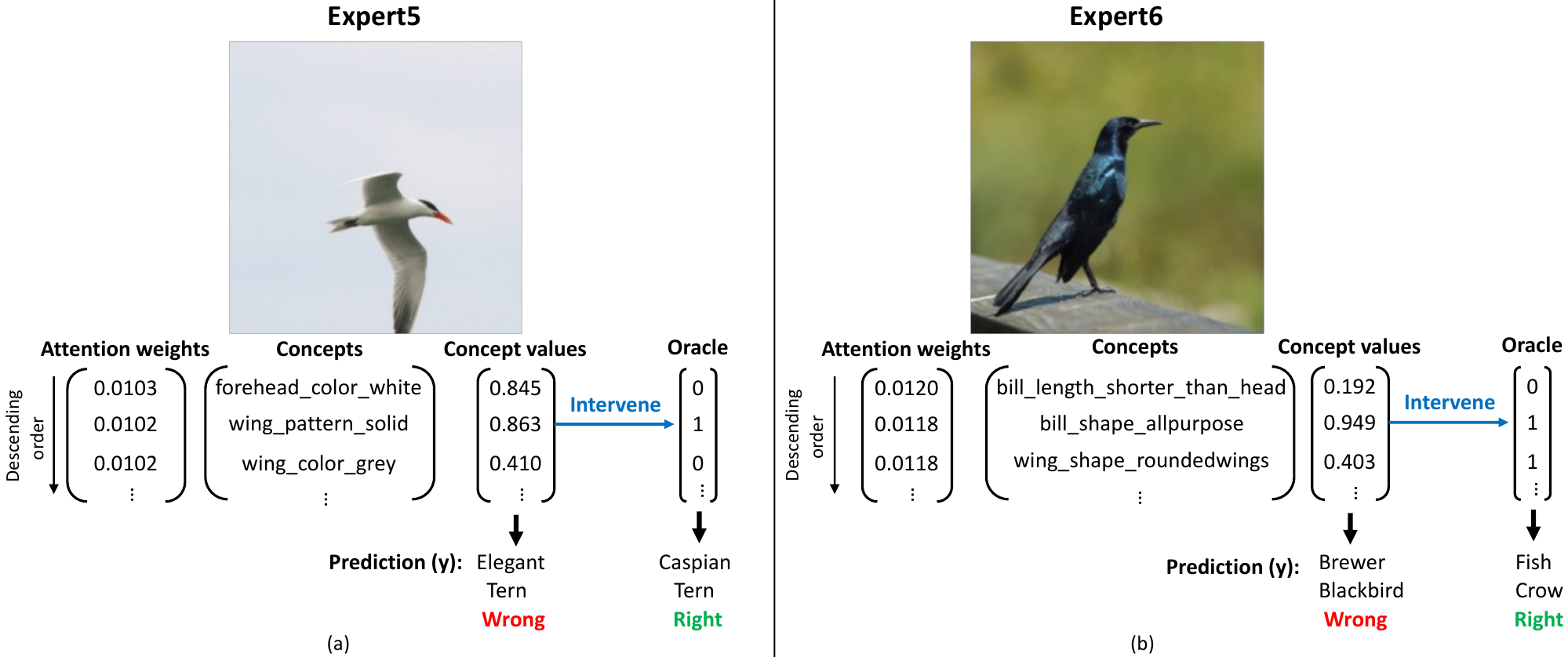}
\caption{Illustration of test time intervention of top-3 concepts for ``harder'' samples identified by the last two experts of VIT-driven MoIE. We adopt E-LEN~\cite{barbiero2022entropy} as the experts. Thus, each concept is associated with an attention weight after training, signifying its prediction importance. So, here we intervene the top 3 concepts with the highest attention weights for samples routed to expert 5 (\textbf{\emph{left}}) and expert 6 (\textbf{\emph{right}}). These samples are considered the ``harder'' samples as they are routed to the last two experts of MoIE. We demonstrate that the test time intervention corrects the prediction. }
\label{fig:tti_qual}
\vspace{-2.5pt}
\end{figure}
~\cref{fig:tti_qual} demonstrates an example of test time intervention of concepts for ``harder'' samples identified by the last two experts of VIT-driven MoIE.

\subsubsection{Diversity of explanations for CUB}
\label{app:local_cub}
\begin{figure}[t]
\centering
\includegraphics[width=1 \linewidth]{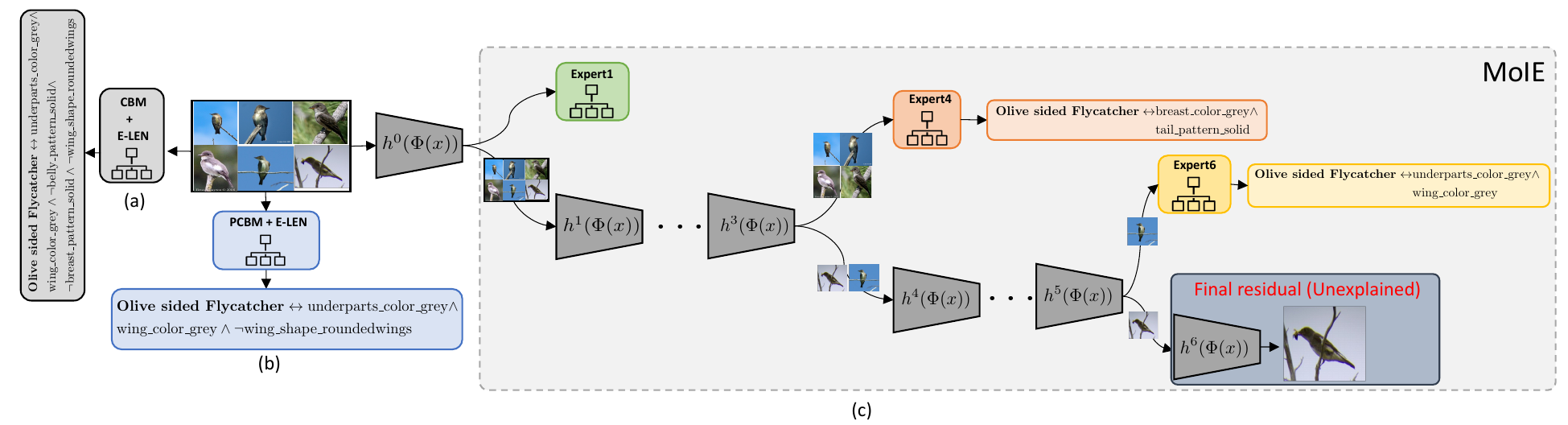}
\vspace{-10pt}
\caption{Construction logical explanations of the samples of a category, ``Olive sided Flycatcher'' in the CUB-200 dataset for (a) VIT-based sequential CBM + E-LEN as an \emph{interpretable by design} baseline, (b) VIT-based PCBM + E-LEN as a posthoc based baseline, (c) various experts in MoIE at inference. This is an example where the final residual covers the unexplained sample, which is ``harder'' to explain (indicated in \emph{red}). Also, MoIE can capture more instance-specific concepts than generic ones by the baselines.}
\label{fig:local_ex_cub_olive_sided}
\vspace{-2.5pt}
\end{figure}

\begin{figure}[h]
\centering
\includegraphics[width=1 \linewidth]{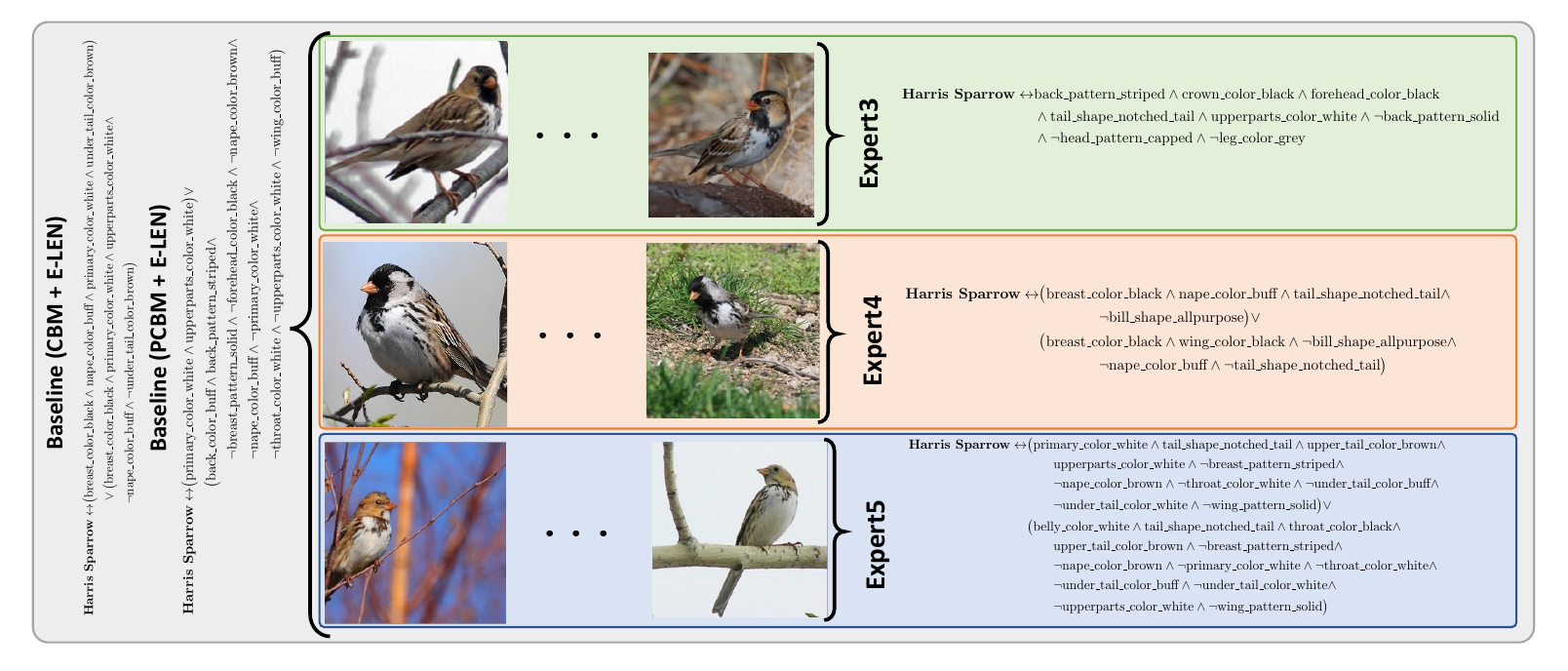}
\vspace{-10pt}
\caption{Construction logical explanations of the samples of a category, ``Harris Sparrow'' in the CUB-200 dataset for (a) VIT-based sequential CBM + E-LEN as an \emph{interpretable by design} baseline, (b) VIT-based PCBM + E-LEN as a posthoc based baseline, (c) various experts in MoIE at inference.}
\label{fig:local_ex_cub_harris}
\vspace{-2.5pt}
\end{figure}

\begin{figure}[h]
\centering
\includegraphics[width=1 \linewidth]{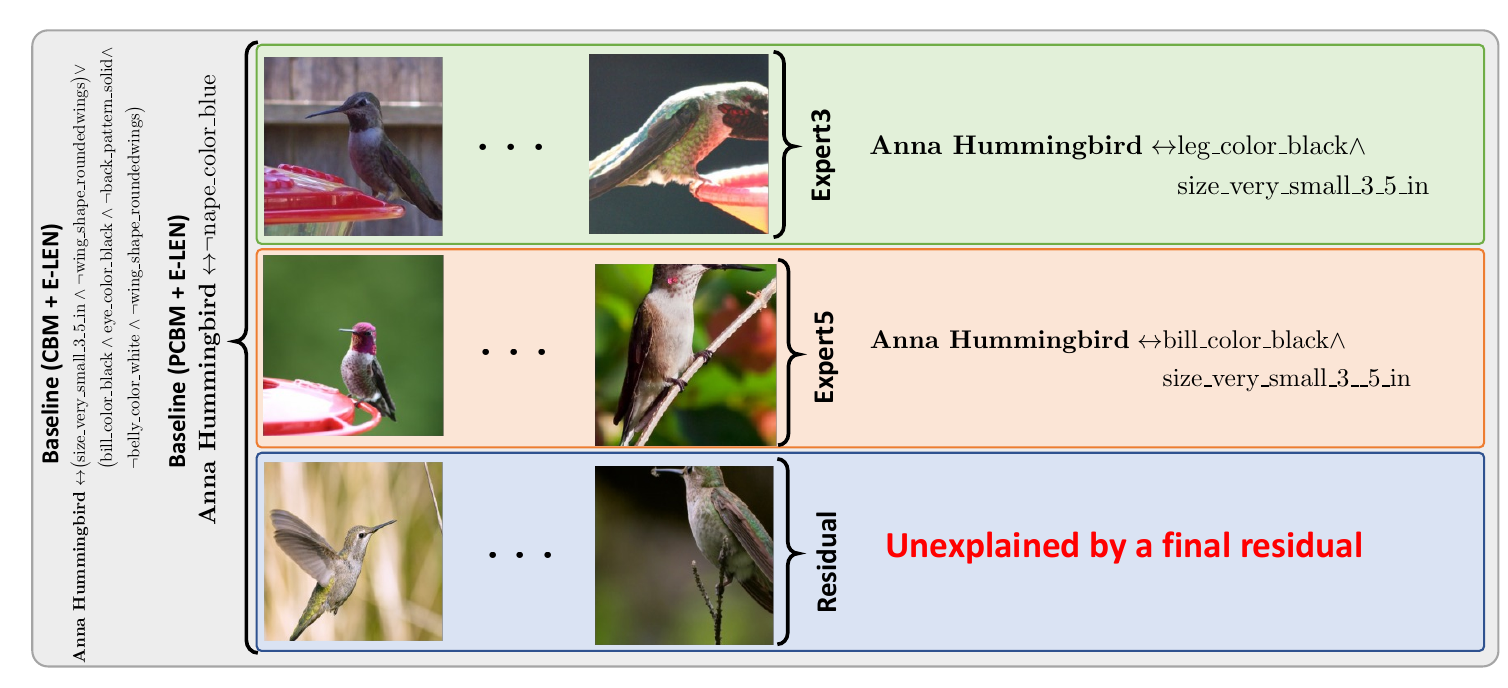}
\vspace{-10pt}
\caption{Construction logical explanations of the samples of a category, ``Anna Hummingbird'' in the CUB-200 dataset for (a) VIT-based sequential CBM + E-LEN as an \emph{interpretable by design} baseline, (b) VIT-based PCBM + E-LEN as a posthoc based baseline, (c) various experts in MoIE at inference.}
\label{fig:local_ex_anna}
\vspace{-2.5pt}
\end{figure}

\begin{figure}[h]
\centering
\includegraphics[width=1 \linewidth]{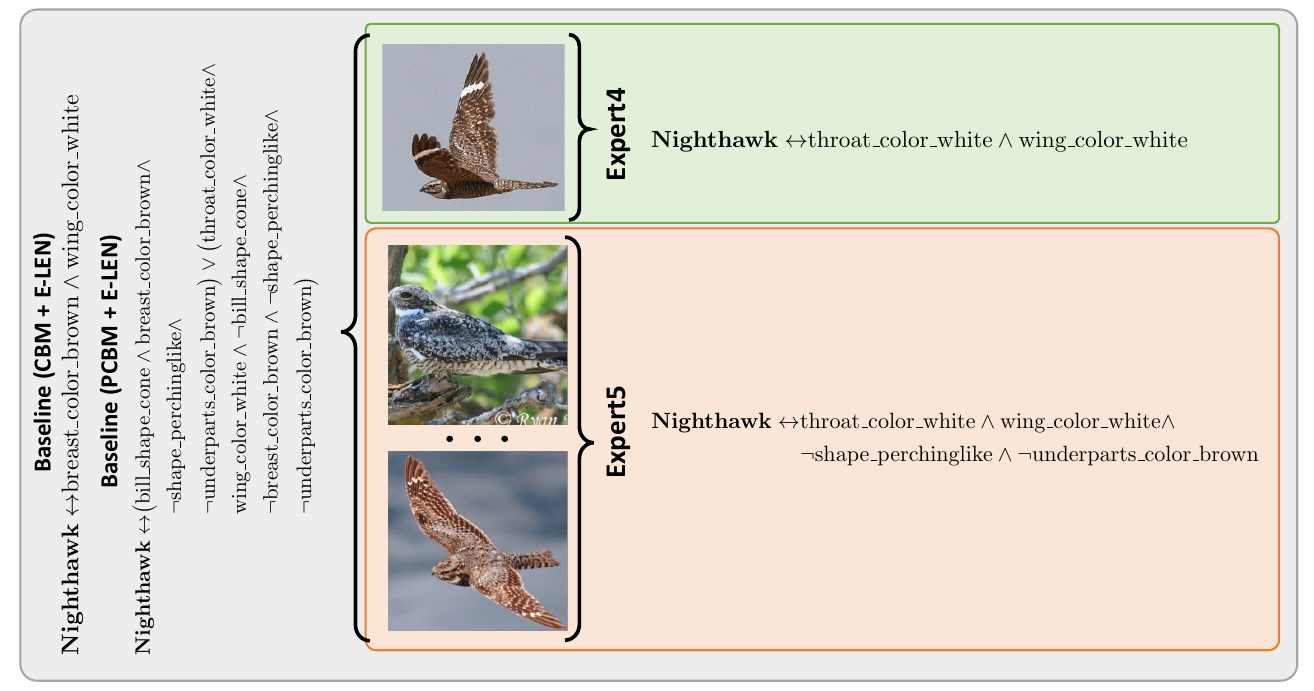}
\vspace{-10pt}
\caption{Construction logical explanations of the samples of a category, ``Nighthawk'' in the CUB-200 dataset for (a) VIT-based sequential CBM + E-LEN as an \emph{interpretable by design} baseline, (b) VIT-based PCBM + E-LEN as a posthoc based baseline, (c) various experts in MoIE at inference. }
\label{fig:local_ex_nighthawk}
\vspace{-2.5pt}
\end{figure}

\begin{figure}[h]
\centering
\includegraphics[width=1 \linewidth]{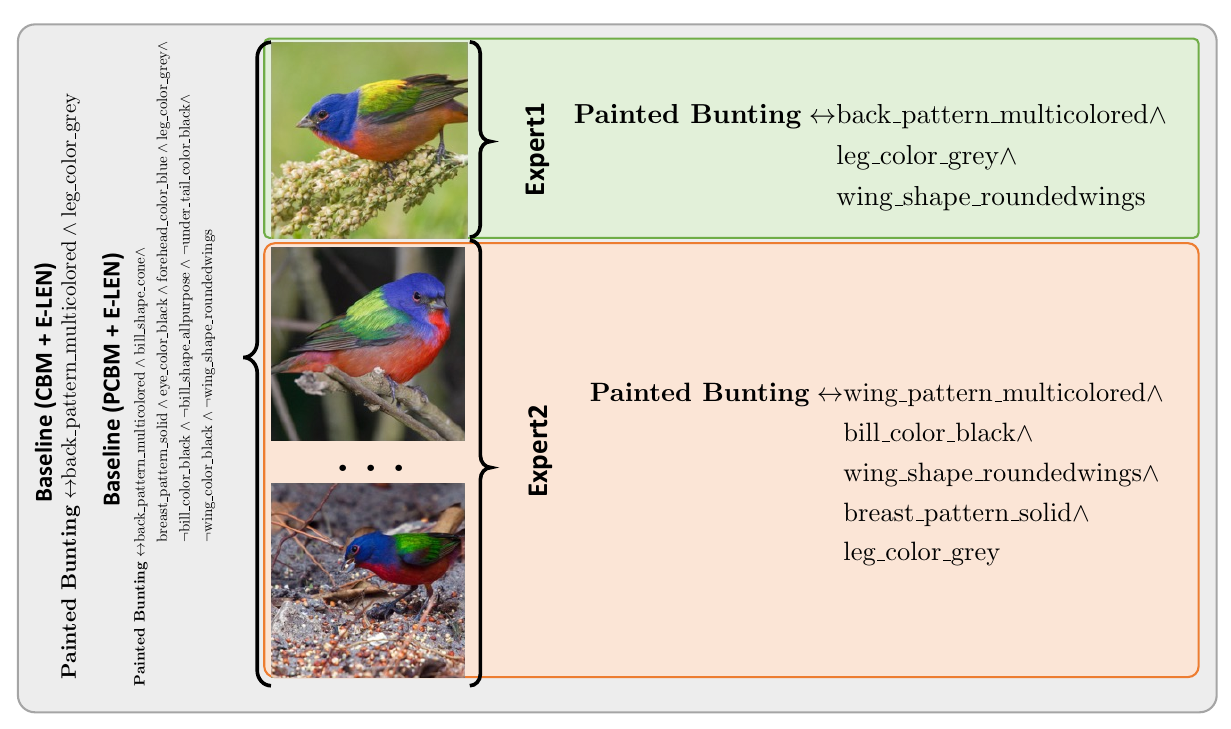}
\vspace{-10pt}
\caption{Construction logical explanations of the samples of a category, ``Painted Bunting'' in the CUB-200 dataset for (a) VIT-based sequential CBM + E-LEN as an \emph{interpretable by design} baseline, (b) VIT-based PCBM + E-LEN as a posthoc based baseline, (c) various experts in MoIE at inference. }
\label{fig:local_ex_cub_painted}
\vspace{-2.5pt}
\end{figure}

\cref{fig:local_ex_cub_olive_sided} shows the construction of instance-specific local FOL explanations of a category, ``Olive sided Flycatcher'' in the CUB-200 dataset for the VIT-based baselines and MoIE. In this example, the final expert6 covers the relatively ``harder'' sample.~\cref{fig:local_ex_cub_harris},~\cref{fig:local_ex_anna},~\cref{fig:local_ex_nighthawk},~\cref{fig:local_ex_cub_painted} shows more such FOL explanations. All these examples demonstrate MoIE's high capability to identify more meaningful instance-specific concepts in FOL explanations. In contrast, the baselines identify the generic concepts for all samples in a class.

\subsubsection{Diversity of explanations for Awa2}
\label{app:local_awa2}
\begin{figure*}[h]
\centering
\includegraphics[width=\columnwidth]{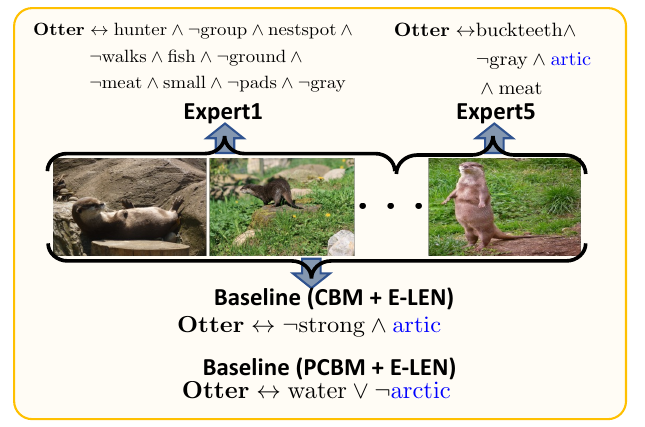}
\vspace{-10pt}
\caption{Flexibility of FOL explanations by VIT-derived MoIE  MoIE and the CBM + E-LEN and PCBM + E-LEN baselines for Awa2 dataset to classify ``Otter'' at inference. Both the baseline's FOL constitutes identical concepts to distinguish all the samples. However, expert1 classifies ``Otter'' with \textit{hunter}, \textit{group} \etc as the identifying concept for the instances covered by it. Similarly expert5 classifies ``Otter'' using \textit{buckteeth}, \textit{gray} \etc. Note that, \textit{meat} and \textit{gray}  are shared between the two experts. We highlight the shared concepts (\textit{artic}) between the experts and the baselines as blue.}
\label{fig:local_awa2_otter}
\vspace{-2.5pt}
\end{figure*}

\begin{figure*}[h]
\centering
\includegraphics[width=20cm,
  height=20cm,
  keepaspectratio]{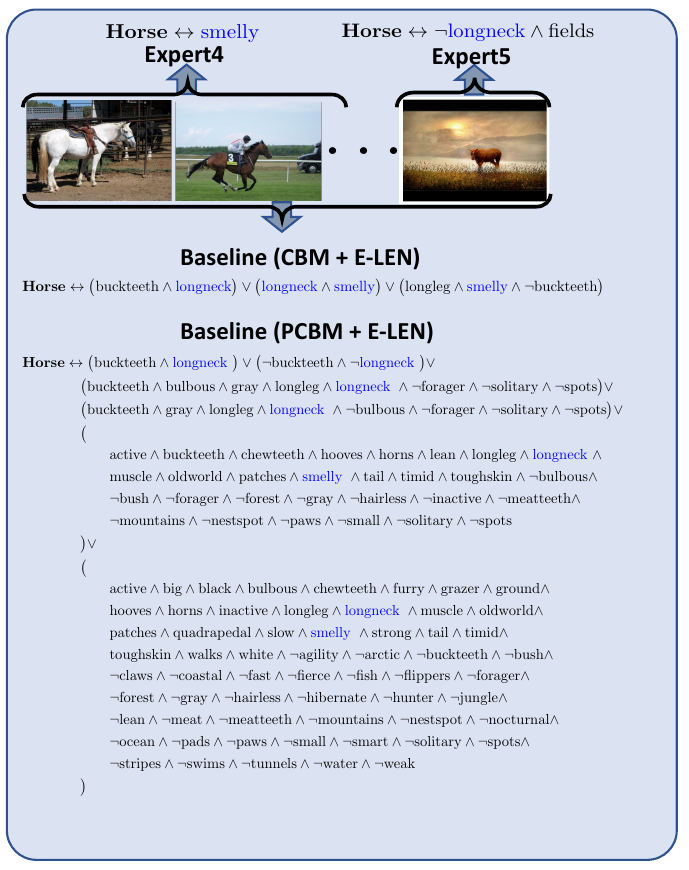}
\vspace{-10pt}
\caption{Flexibility of FOL explanations by VIT-derived MoIE  MoIE and the CBM + E-LEN and PCBM + E-LEN baselines for Awa2 dataset to classify ``Horse'' at inference. Both the baseline's FOL constitutes identical concepts to distinguish all the samples. However, expert4 classifies ``Horse'' with \textit{smelly} as the identifying concept for the instances covered by it. Similarly, expert5 classifies the same ``Horse'' using \textit{longneck} and \textit{fields}. We highlight the shared concepts between the experts and the baselines as blue.}
\label{fig:local_awa2_horse}
\vspace{-2.5pt}
\end{figure*}

\cref{fig:local_awa2_otter} and~\ref{fig:local_awa2_horse} demonstrate the flexibility of instance-specific local FOL explanations by VIT-derived MoIE compared to the different baselines for the Awa2 dataset qualitatively.

\subsubsection{VIT-based experts compose of less concepts than the ResNet-based counterparts}
\label{app:comparison_arch}
\begin{figure}[h]
\centering
\includegraphics[width=1\textwidth]{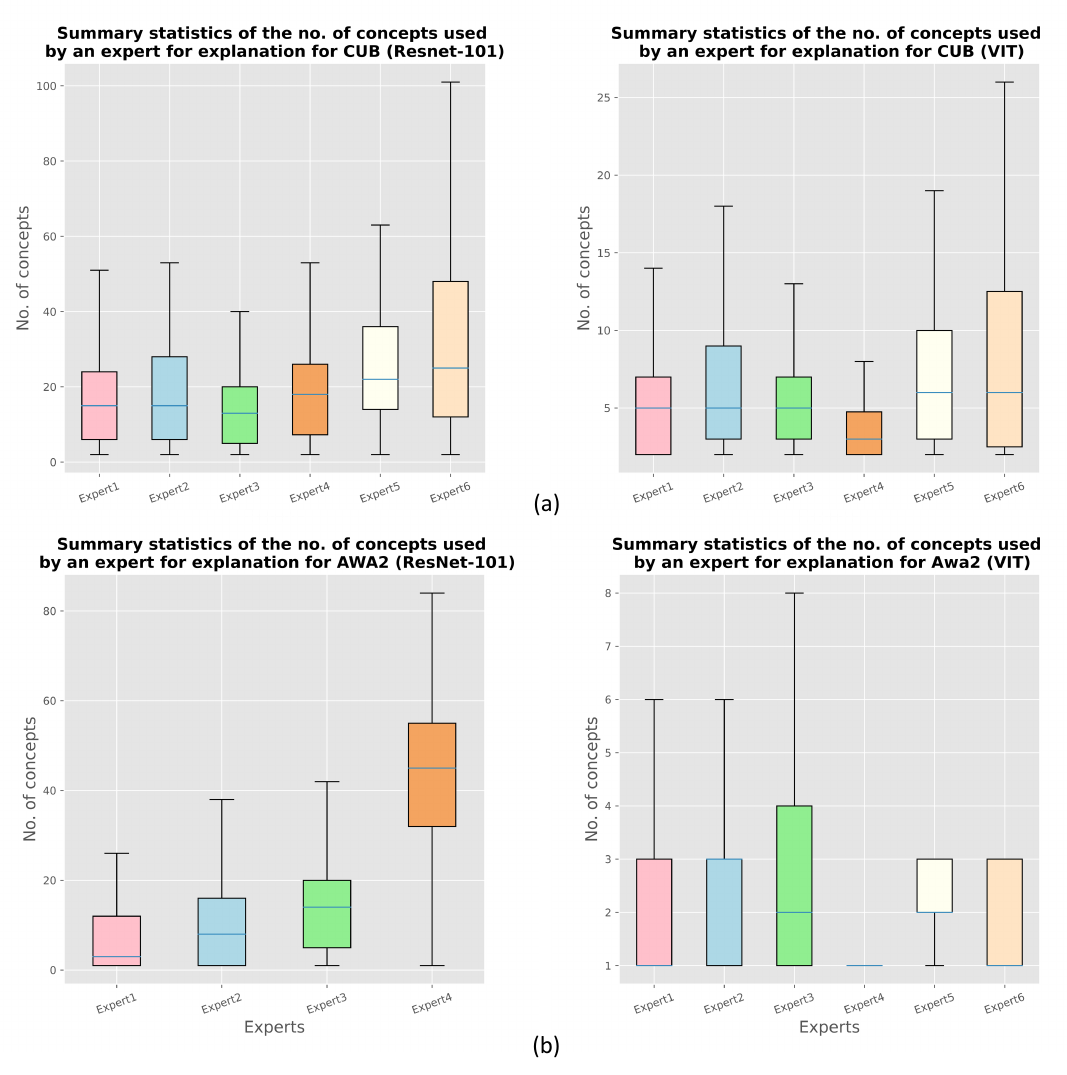}
\caption{Summary statistics of the number of concepts utilized by various experts of datasets (a) CUB -200(top row) and (b) Awa2 (bottom row). In general, we can see that experts carving out the explanations from VIT often uses less number of concepts.}
\label{fig:stats_ex_cub}
\end{figure}

\begin{figure}
\centering
\includegraphics[width=14cm, height=13cm]
{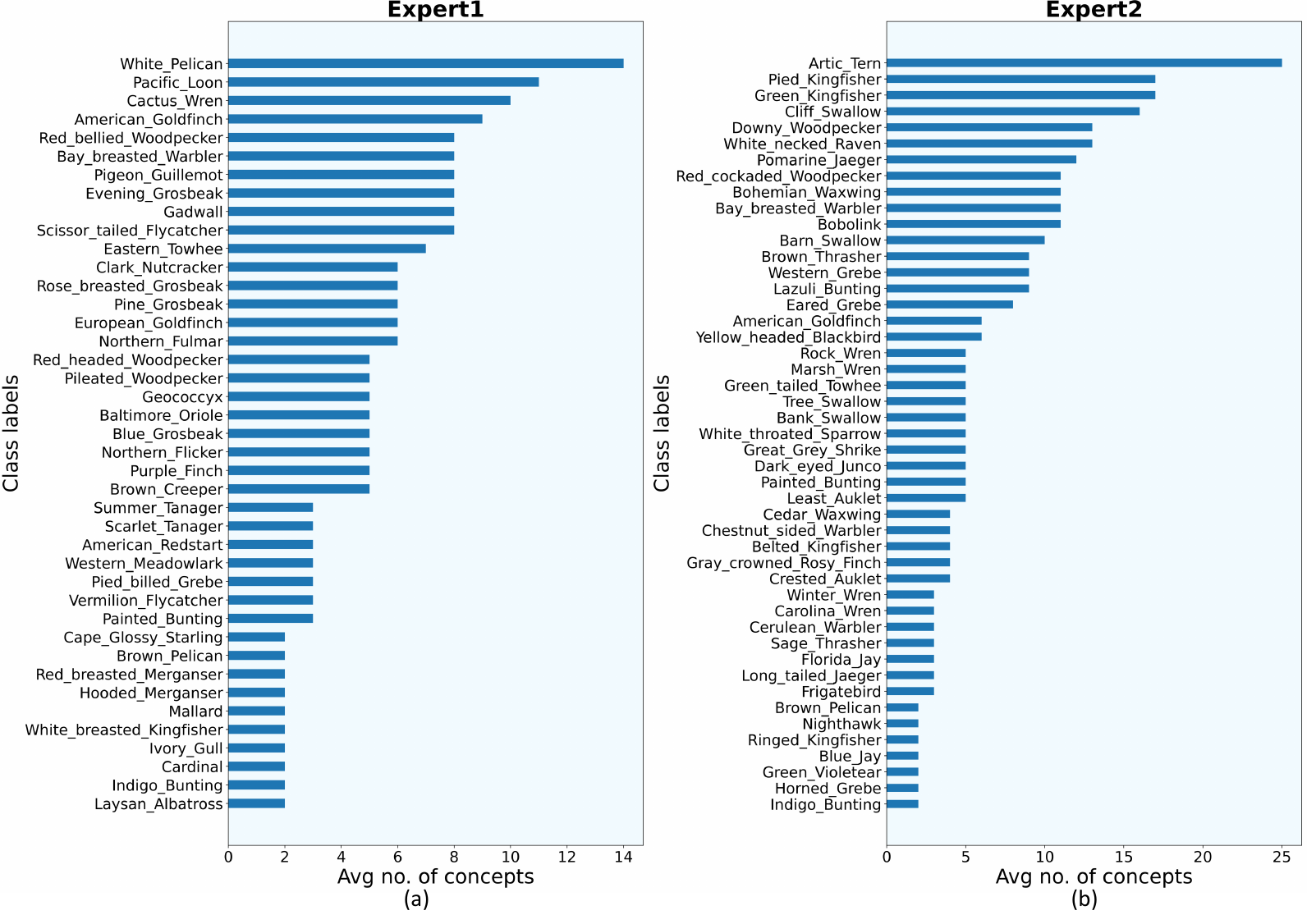}
\caption{Class labels (Bird species) vs. avg concepts using VIT as the backbone for CUB-200 by (a) Expert1 (b) Expert2. Each bar in this plot indicates the average number of concepts required to explain each sample of that bird species correctly. For example according to (a) expert1 requires 14 concepts to explain an instance of ``White Pelican''.}
\label{fig:vit_cub_concept_1_2}
\end{figure}

\begin{figure}
\centering
\includegraphics[width=14cm, height=13cm]
{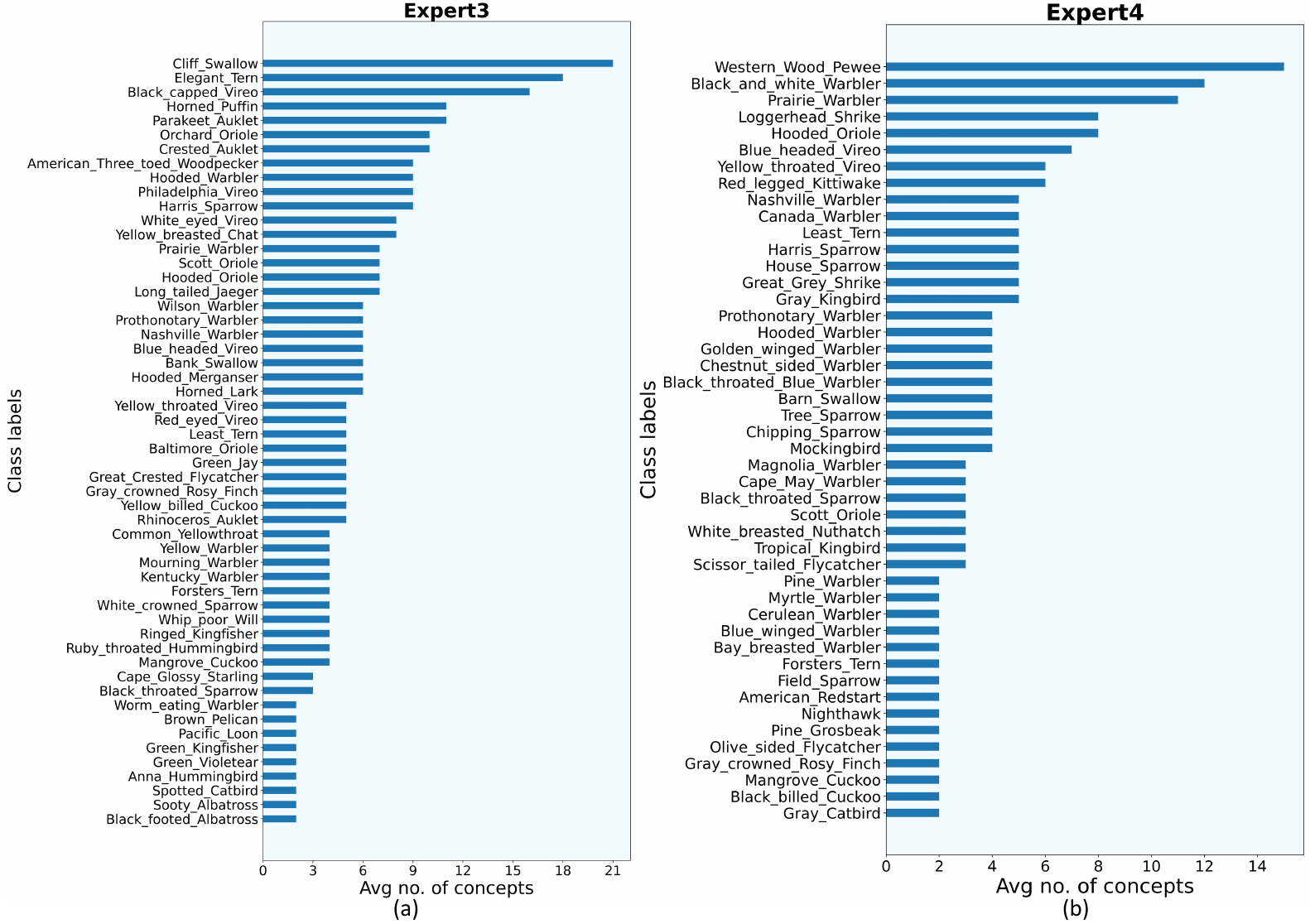}
\caption{Class labels (Bird species) vs. avg concepts using VIT as the backbone for CUB-200 by (a) Expert3 (b) Expert4. Each bar in this plot indicates the average number of concepts required to explain each sample of that bird species correctly. For example according to (a) expert3 requires 21 concepts to explain an instance of ``Cliff Swallow''.}
\label{fig:vit_cub_concept_3_4}
\end{figure}

\begin{figure}
\centering
\includegraphics[width=14cm, height=13cm]
{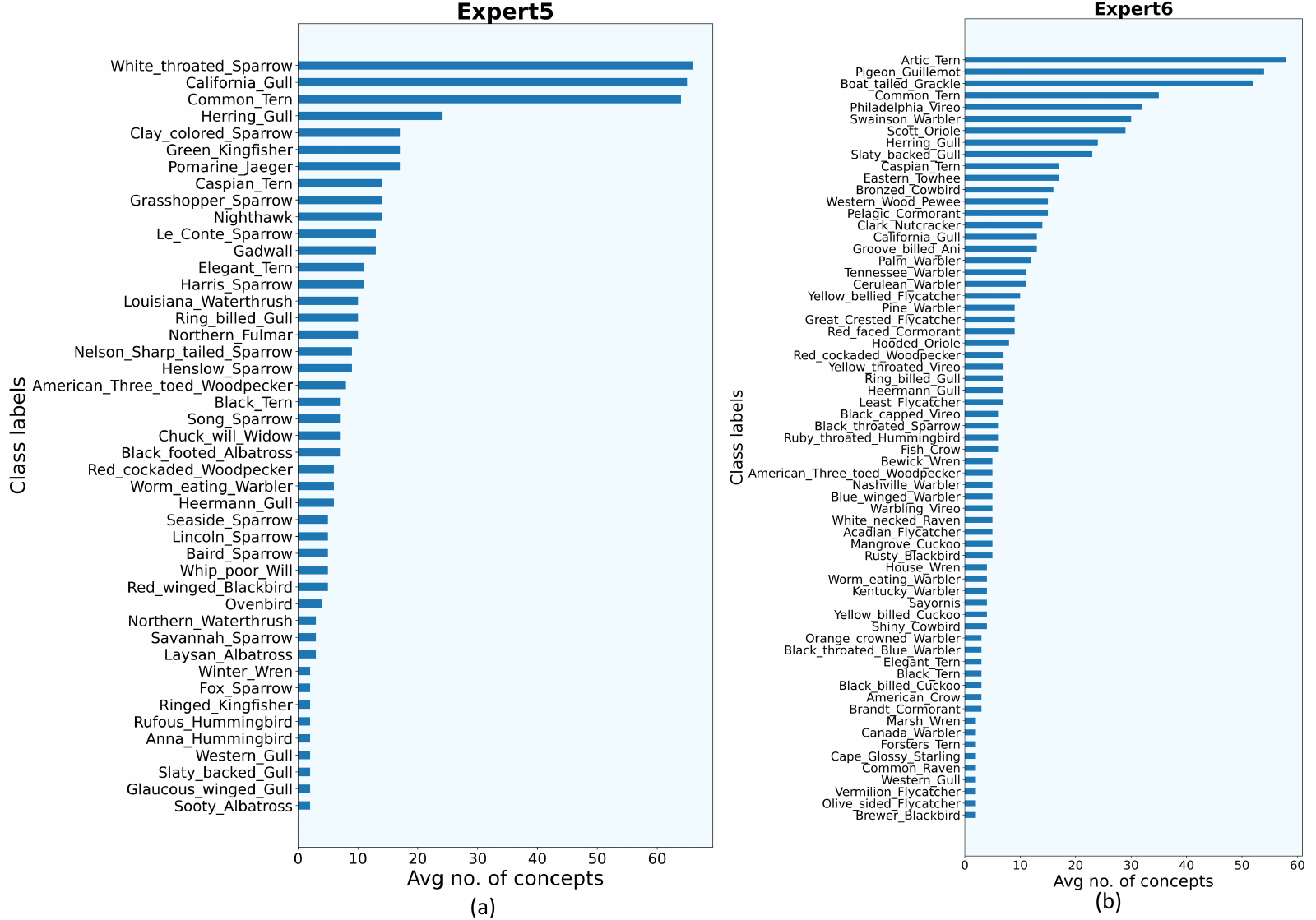}
\caption{Class labels (Bird species) vs. avg concepts using VIT as the backbone for CUB-200 by (a) Expert5 (b) Expert6. Each bar in this plot indicates the average number of concepts required to explain each sample of that bird species correctly. For example according to (a) expert5 requires approximately 65 concepts to explain an instance of ``White throated sparrow''.}
\label{fig:vit_cub_concept_5_6}
\end{figure}

\begin{figure}
\centering
\includegraphics[width=15cm, height=13cm]
{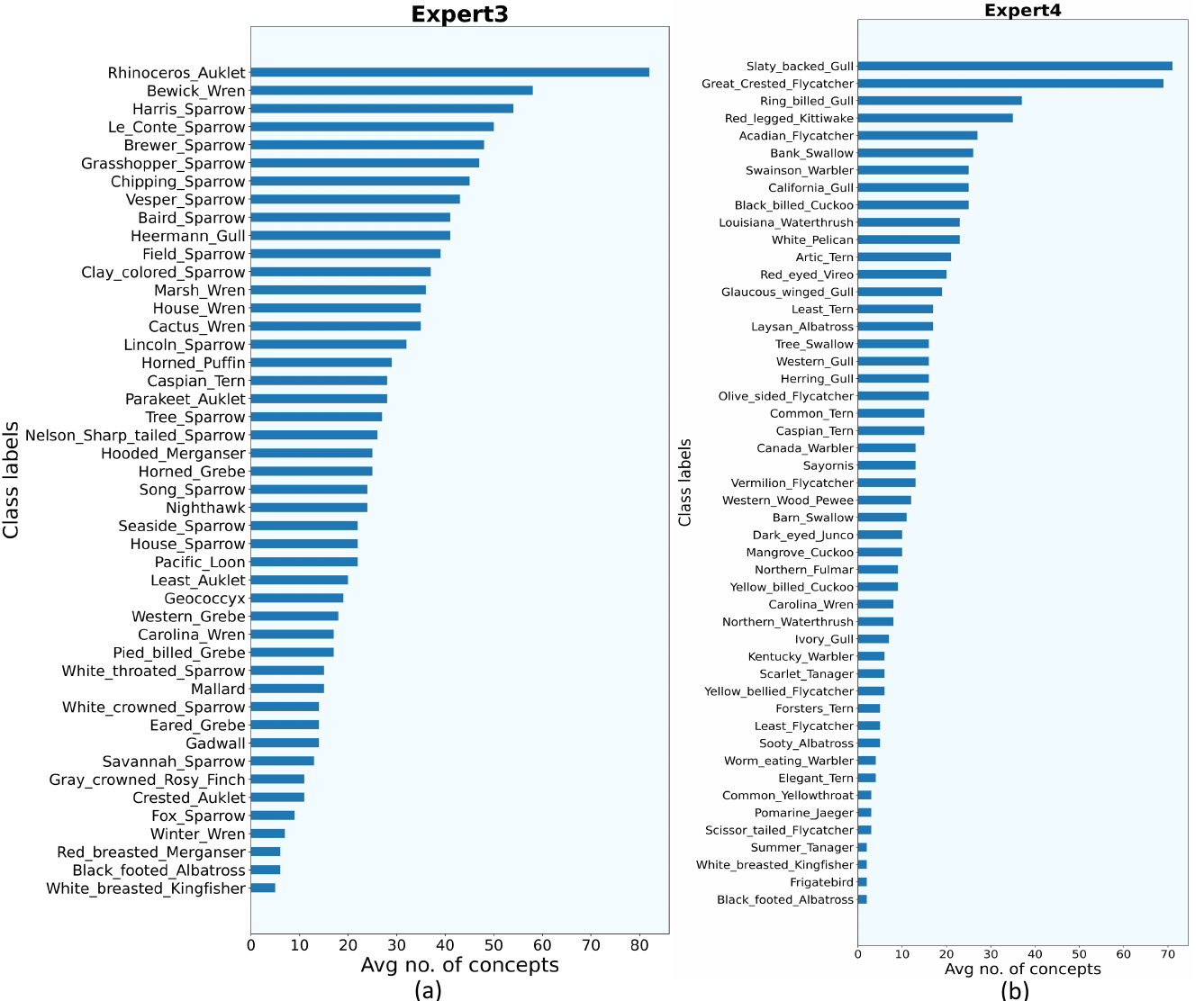}
\caption{Class labels (Bird species) vs. avg concepts using  ResNet-101 as the backbone for CUB-200 by (a) Expert3 (b) Expert4. Each bar in this plot indicates the average number of concepts required to explain each sample of that bird species correctly. For example according to (a) expert3 requires approximately 82 concepts to explain an instance of ``Rhinoceros auklet''.}
\label{fig:cnn_cub_concept_3_4}
\end{figure}

\cref{fig:stats_ex_cub} shows the summary statistics for multiclass classification vision datasets. For both datasets, we observe that the VIT-based MoIE uses fewer concepts for explanation than their ResNet-based counterparts. For example, for the CUB-200 dataset, expert6 of VIT-backbone requires 25 concepts compared to 105 by expert6 of ResNet-101-backbone (~\cref{fig:stats_ex_cub}a). The 105 concepts by expert6 is the highest number of concepts utilized by any expert for CUB-200. Similarly, for Awa2, the highest number concept used by an expert is 8 for the VIT-based backbone compared to 80 for the ResNet-101-based backbone (\cref{fig:stats_ex_cub}b).
As mentioned before, the average number of concepts for class $j$ = $\frac{\sum\text{all concepts for the samples belong to class $j$}}{\text{\# samples of class $j$}}$. We can see that for ResNet-101, on average 80 concepts are required to explain a sample correctly for the class ``Rhinoceros\_Auklet'' (expert3 in~\cref{fig:cnn_cub_concept_3_4} a). However, for VIT, only 6 concepts are needed to explain a sample correctly ``Rhinoceros\_Auklet'' (expert3  in~\cref{fig:vit_cub_concept_3_4} a). From both of these figures, we can see that different experts require a different number of concepts to explain the same class. For example,~\cref{fig:vit_cub_concept_1_2}  (b) and~\cref{fig:vit_cub_concept_5_6} (b) reveal that experts 2 and 6 require 25 and 58 concepts on average to explain ``Artic\_Tern'' correctly respectively for VIT-derived MoIE.

\begin{figure}
\centering
\includegraphics[width=1\linewidth]
{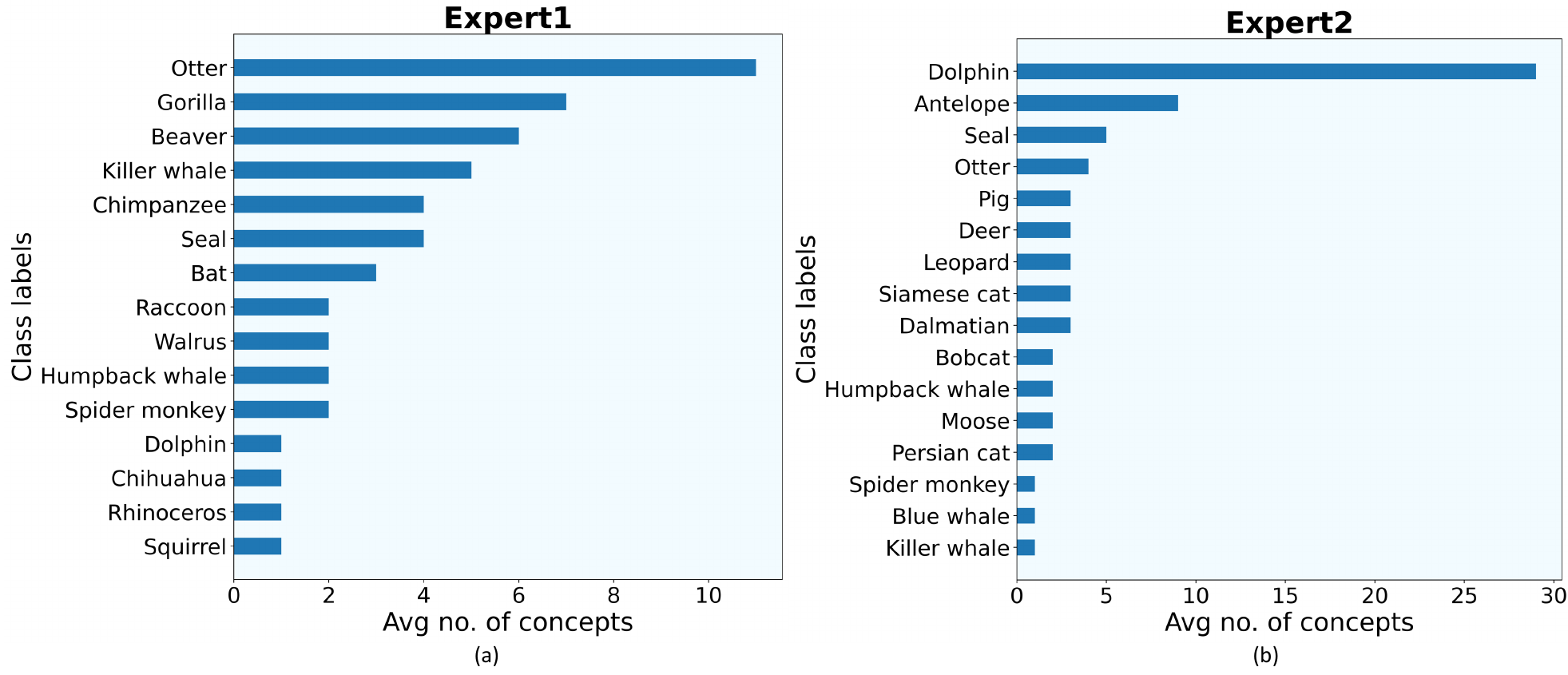}
\caption{Class labels (Animal species) vs. avg concepts using VIT as the backbone for Awa2. Each bar in this plot indicates the average number of concepts required to explain each sample of that animal species correctly. For example according to (c) expert1 requires approximately 12 concepts to explain an instance of ``Otter''.}
\label{fig:Awa2_VIT_a}
\end{figure}

\begin{figure}
\centering
\includegraphics[width=1\linewidth]
{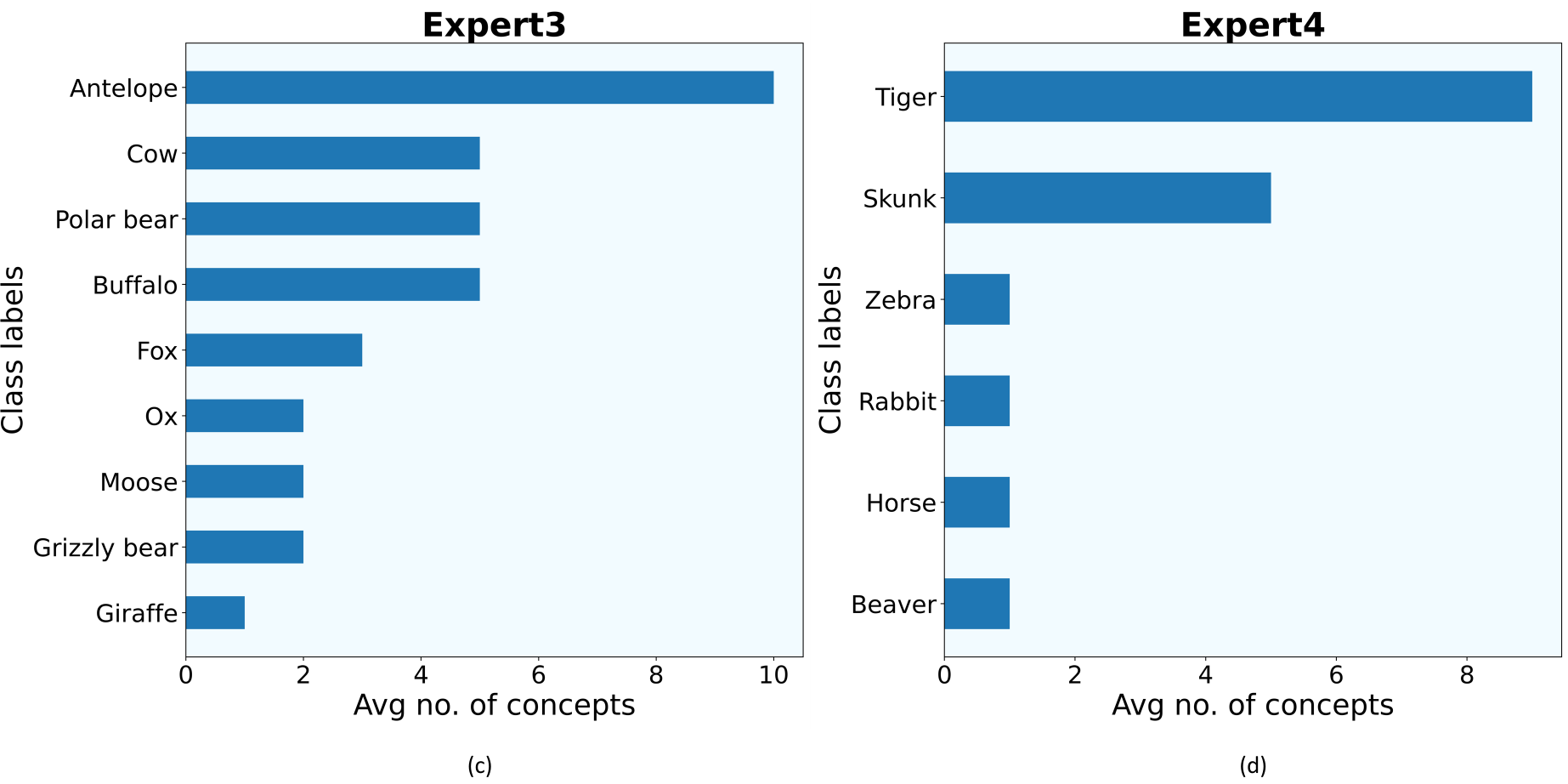}
\caption{Class labels (Animal species) vs. avg concepts using VIT as the backbone for Awa2. Each bar in this plot indicates the average number of concepts required to explain each sample of that animal species correctly. For example according to (c) expert3 requires approximately 10 concepts to explain an instance of ``Antelope''.}
\label{fig:Awa2_VIT_b}
\end{figure}

\begin{figure}
\centering
\includegraphics[width=1\linewidth]
{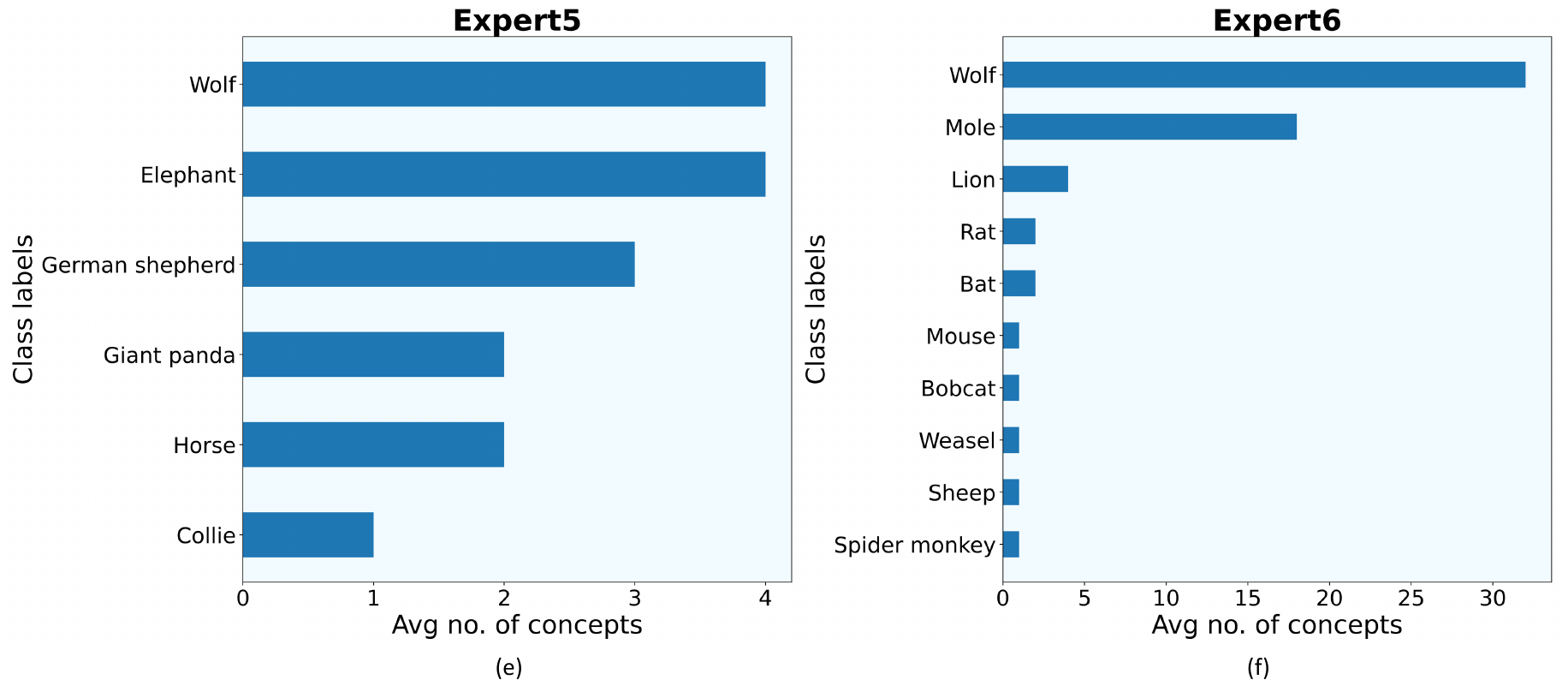}
\caption{Class labels (Animal species) vs. avg concepts using VIT as the backbone for Awa2. Each bar in this plot indicates the average number of concepts required to explain each sample of that animal species correctly. For example according to (e) expert5 requires approximately 4 concepts to explain an instance of ``Antelope''.}
\label{fig:Awa2_VIT_c}
\end{figure}

\begin{figure}[t]
\centering
\includegraphics[width=1\linewidth]
{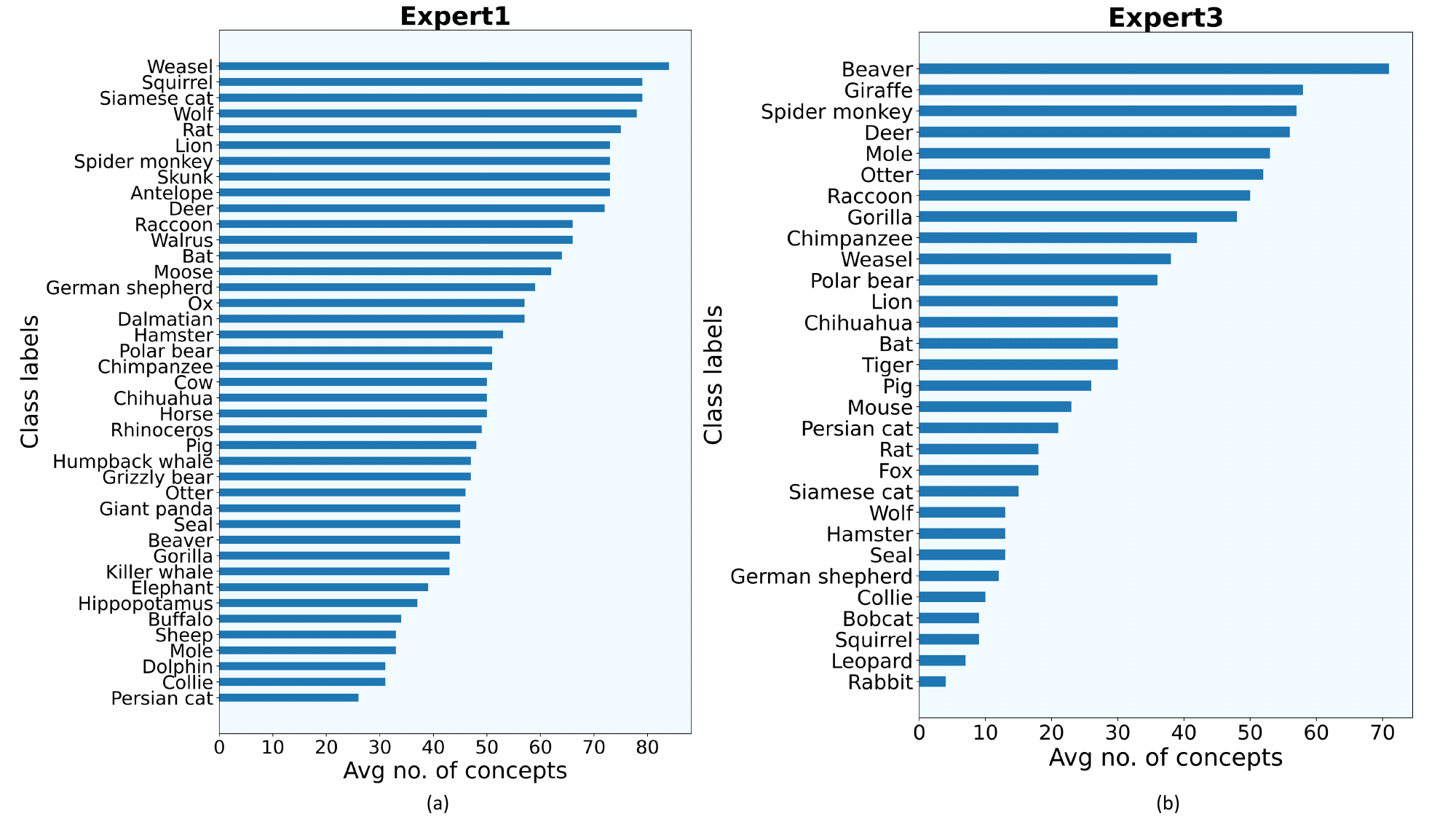}
\caption{Class labels (Animal species) vs. avg concepts using ResNet-101 as the backbone for Awa2. Each bar in this plot indicates the average number of concepts required to explain each sample of that animal species correctly. For example according to (a) expert1 requires approximately 80 concepts to explain an instance of ``Weasel''.}
\label{fig:Awa2_CNN_a}
\end{figure}

\begin{figure}
\centering
\includegraphics[width=1\linewidth]
{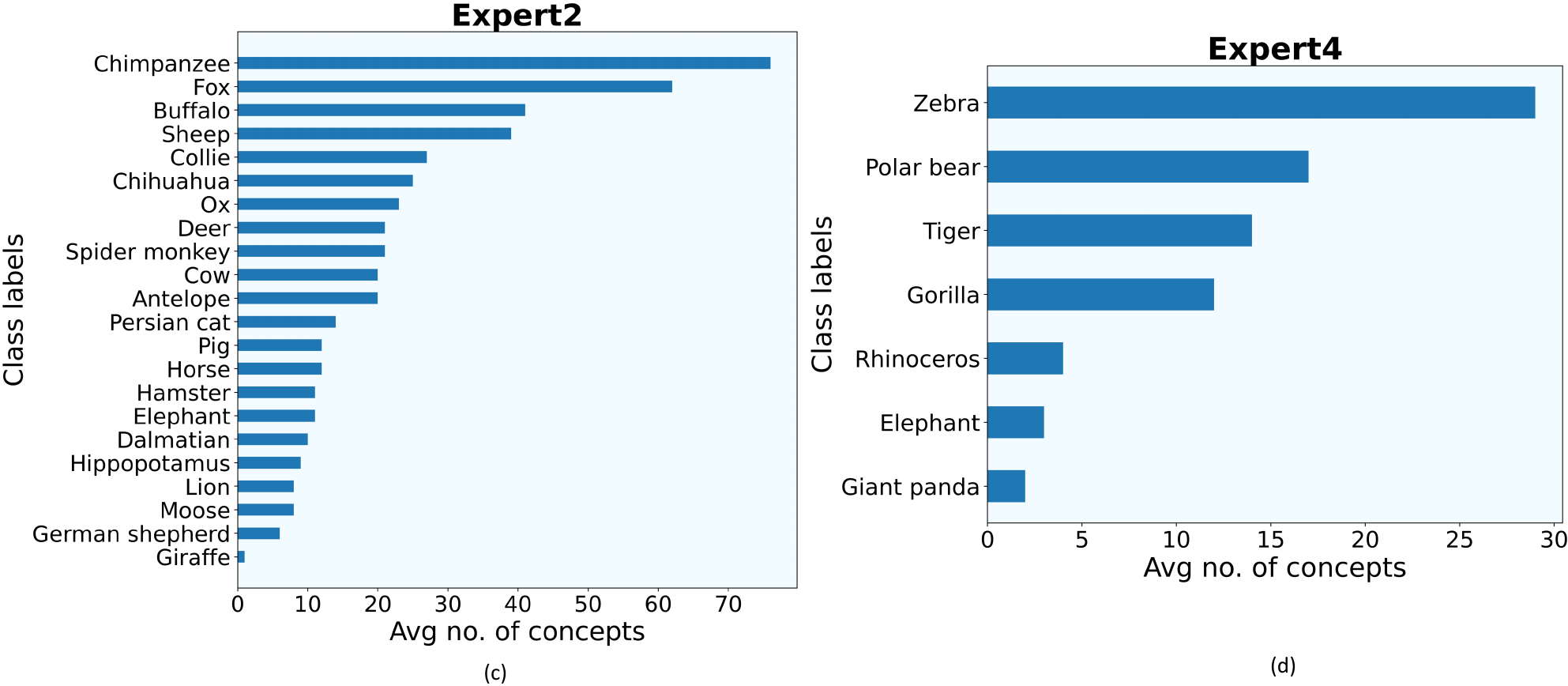}
\caption{Class labels (Animal species) vs. avg concepts using ResNet-101 as the backbone for Awa2. Each bar in this plot indicates the average number of concepts required to explain each sample of that animal species correctly. For example according to (b) expert2 requires approximately 72 concepts to explain an instance of ``Chimpanzee''.}
\label{fig:Awa2_CNN_b}
\end{figure}

~\cref{fig:Awa2_VIT_a}, ~\cref{fig:Awa2_VIT_b},~\cref{fig:Awa2_VIT_c} display the average number of concepts required to predict an animal species correctly in the Awa2 dataset for VIT as backbones. Similarly~\cref{fig:Awa2_CNN_a} and~\cref{fig:Awa2_CNN_b} display the average number of concepts required to predict an animal species correctly in the Awa2 dataset for ResNet101 as backbones. We can see that for ResNet101, on average, 80 concepts are required to explain a sample correctly for the class ``Weasel'' (Expert1 in~\cref{fig:Awa2_CNN_a} a). However, for VIT, only three concepts are needed to explain a sample correctly for ``Weasel'' (Expert 6 in~\cref{fig:Awa2_VIT_c} f). Also, from both of these figures, we can see that different experts require different number concepts to explain the same class. For example,~\cref {fig:Awa2_VIT_c} (e) and (f) reveal that experts 5 and 6 require 4 and 30 concepts on average to explain ``Wolf'' correctly.

\begin{figure}
\centering
\includegraphics[width=15cm, height=13cm]
{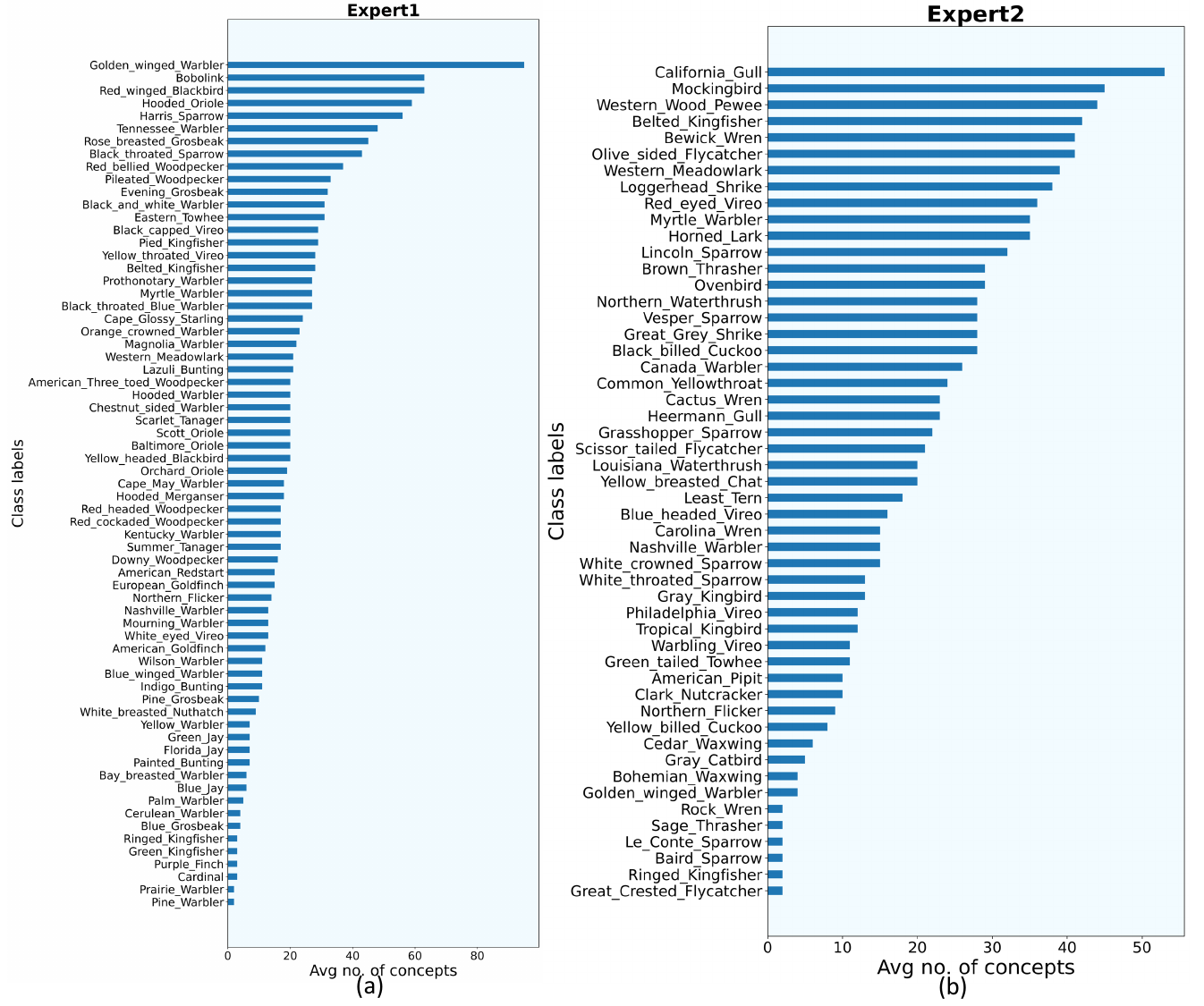}
\caption{Class labels (Bird species) vs. avg concepts using ResNet-101 as the backbone for CUB-200 by (a) Expert1 (b) Expert2. Each bar in this plot indicates the average number of concepts required to explain each sample of that bird species correctly. For example according to (a) expert1 requires approximately 85 concepts to explain an instance of ``Golden winged warbler''.}
\label{fig:cnn_cub_concept_1_2}
\end{figure}

\begin{figure}
\centering
\includegraphics[width=1\linewidth]
{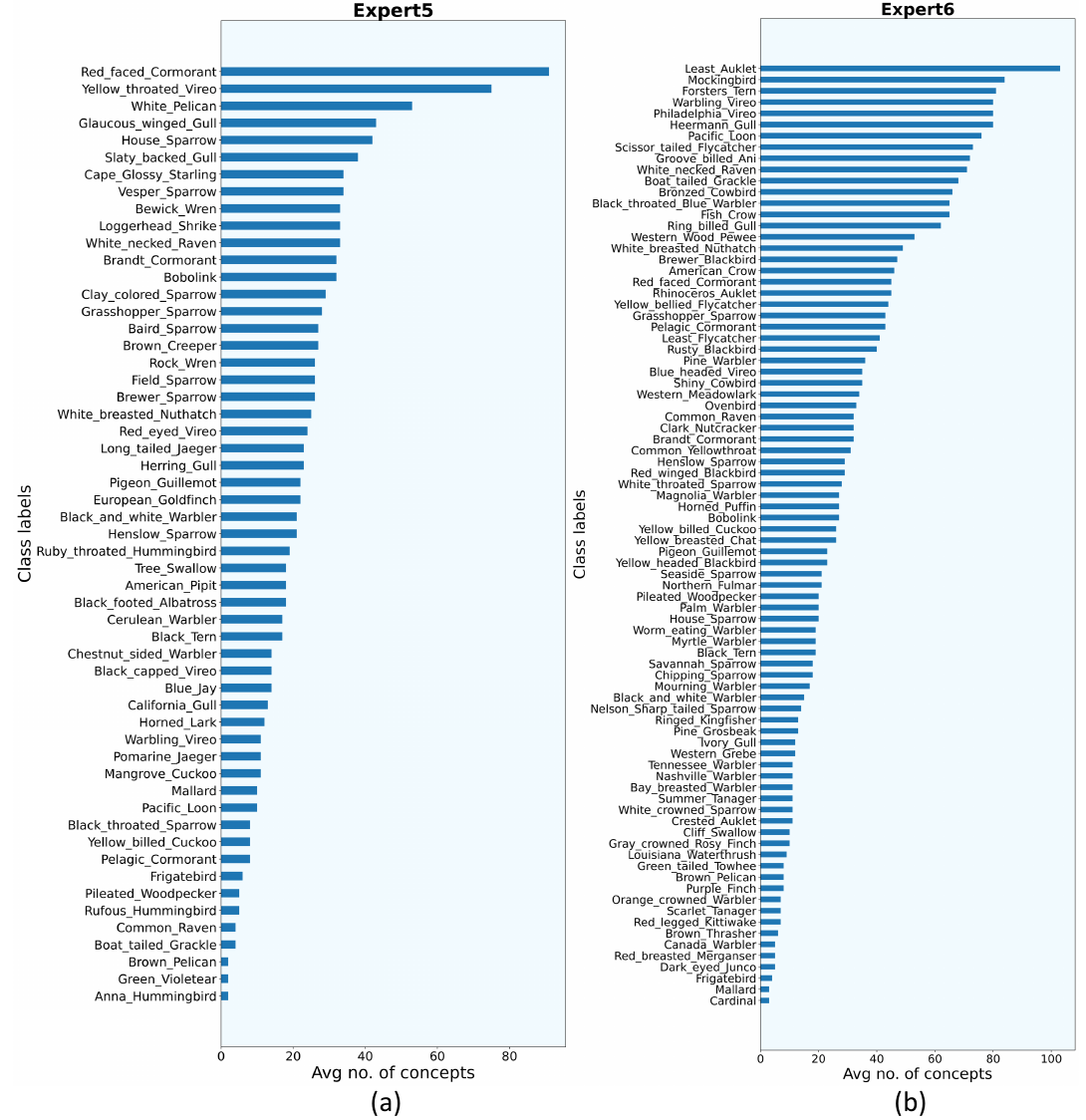}
\caption{Class labels (Bird species) vs. avg concepts using  ResNet-101 as the backbone for CUB-200 by (a) Expert5 (b) Expert6. Each bar in this plot indicates the average number of concepts required to explain each sample of that bird species correctly. For example according to (a) expert5 requires approximately 85 concepts to explain an instance of ``Red faced carmorant''.}
\label{fig:cnn_cub_concept_5_6}
\end{figure}

\subsection{Computational performance }
\label{app:validate_concepts}
\begin{figure*}[h]
\centering
\includegraphics[width=1.0\textwidth]
{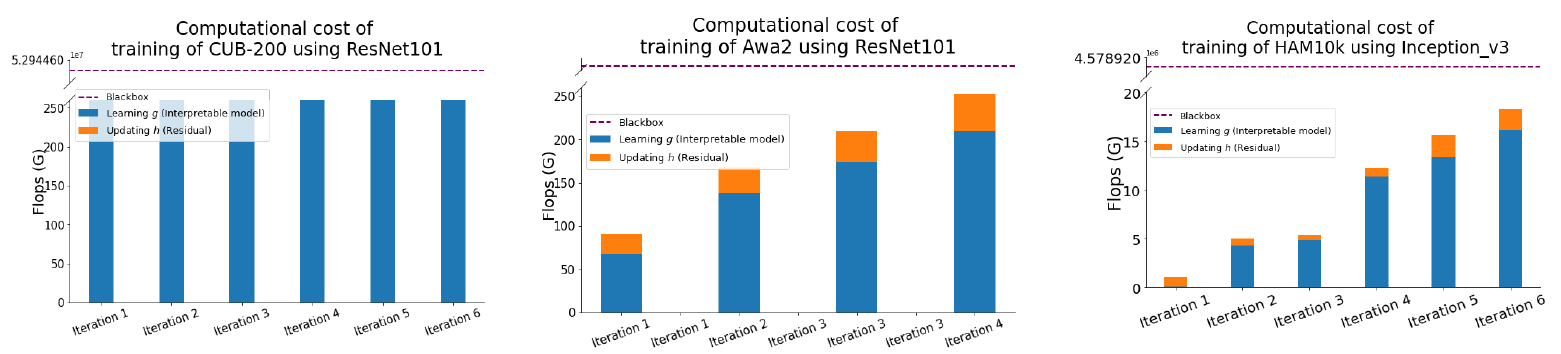}
\caption{Flops vs. iteration for MoIE and the Blackbox. The dotted line in the figure represents the flops taken by the blackbox.}
\label{fig:flops}
\end{figure*}

\cref{fig:flops} shows the computational performance compared to the Blackbox. Though in MoIE, we sequentially learn the experts and the residuals, they take less computational resources than the Blackbox. The experts are shallow neural networks. Also, we only update the classification layer ($h$) for the residuals, so it takes such less time. The Flops in the Y axis are computed as Flop of (forward propagation + backward propagation) $\times$ (minibatch size) $\times$ (no of training epochs).
We use the Pytorch profiler package to monitor the flops.

\end{document}